\documentclass[10pt,twocolumn,letterpaper]{article}

\usepackage[pagenumbers]{cvpr} 

\usepackage[dvipsnames]{xcolor}

\definecolor{cvprblue}{rgb}{0.21,0.49,0.74}
\usepackage[pagebackref,breaklinks,colorlinks,citecolor=cvprblue]{hyperref}
\usepackage{xcolor,colortbl}
\usepackage{multirow}
\usepackage{array,booktabs}
\usepackage{hwemoji}
\usepackage{wasysym}
\usepackage{amsmath}
\usepackage[normalem]{ulem}
\usepackage{gradient-text}
\usepackage{hwemoji}
\usepackage{wasysym}
\usepackage{float}
\usepackage{sidecap}
\usepackage[accsupp]{axessibility}
\DeclareSymbolFont{extraup}{U}{zavm}{m}{n}
\DeclareMathSymbol{\varheart}{\mathalpha}{extraup}{86}
\DeclareMathSymbol{\vardiamond}{\mathalpha}{extraup}{87}
\newcommand{\puicon}{\includegraphics[scale=0.027]{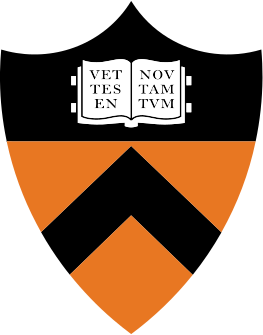}}
\newcommand{\ucsdicon}{\includegraphics[scale=0.007]{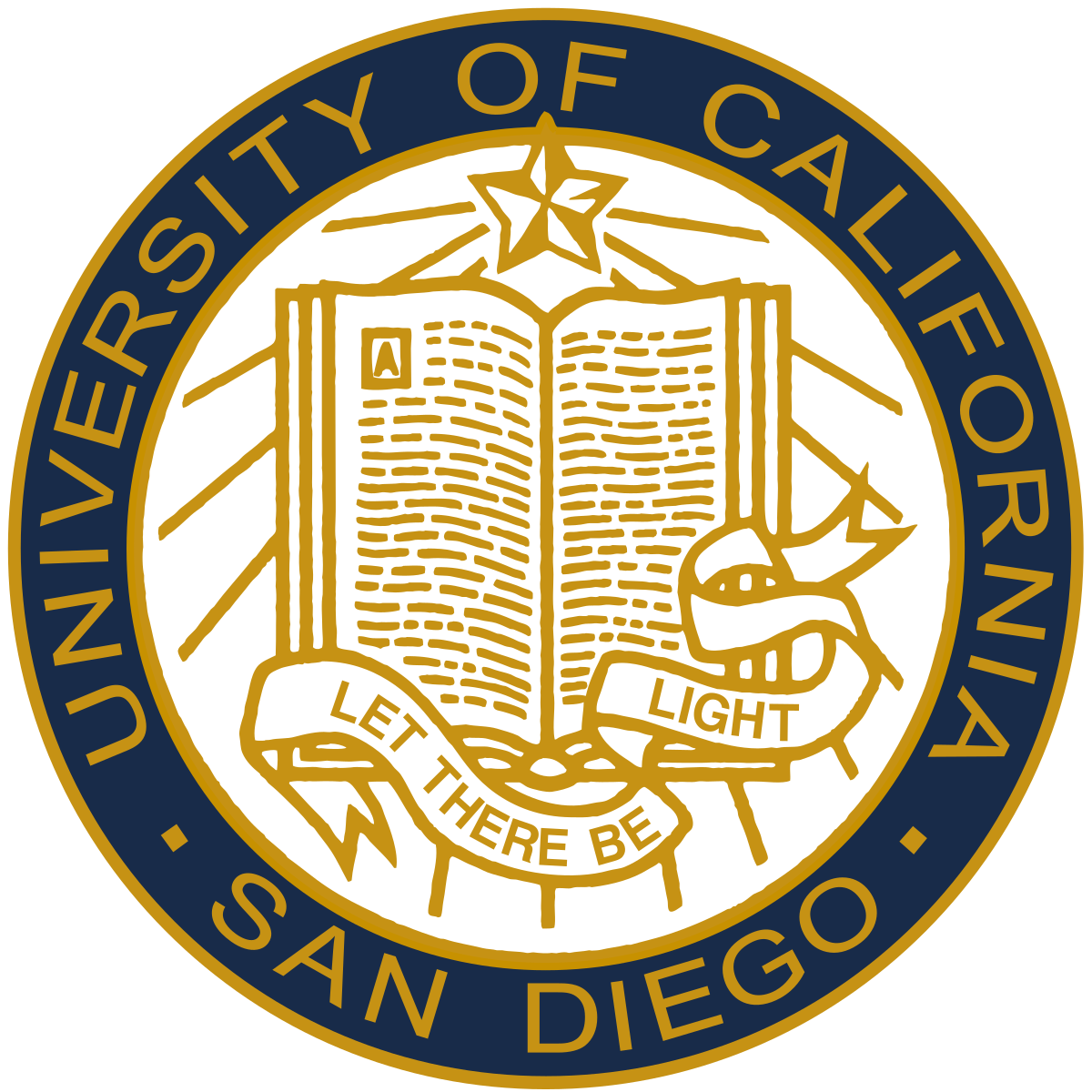}}
\newcommand{\tsinghuaicon}{\includegraphics[scale=0.007]{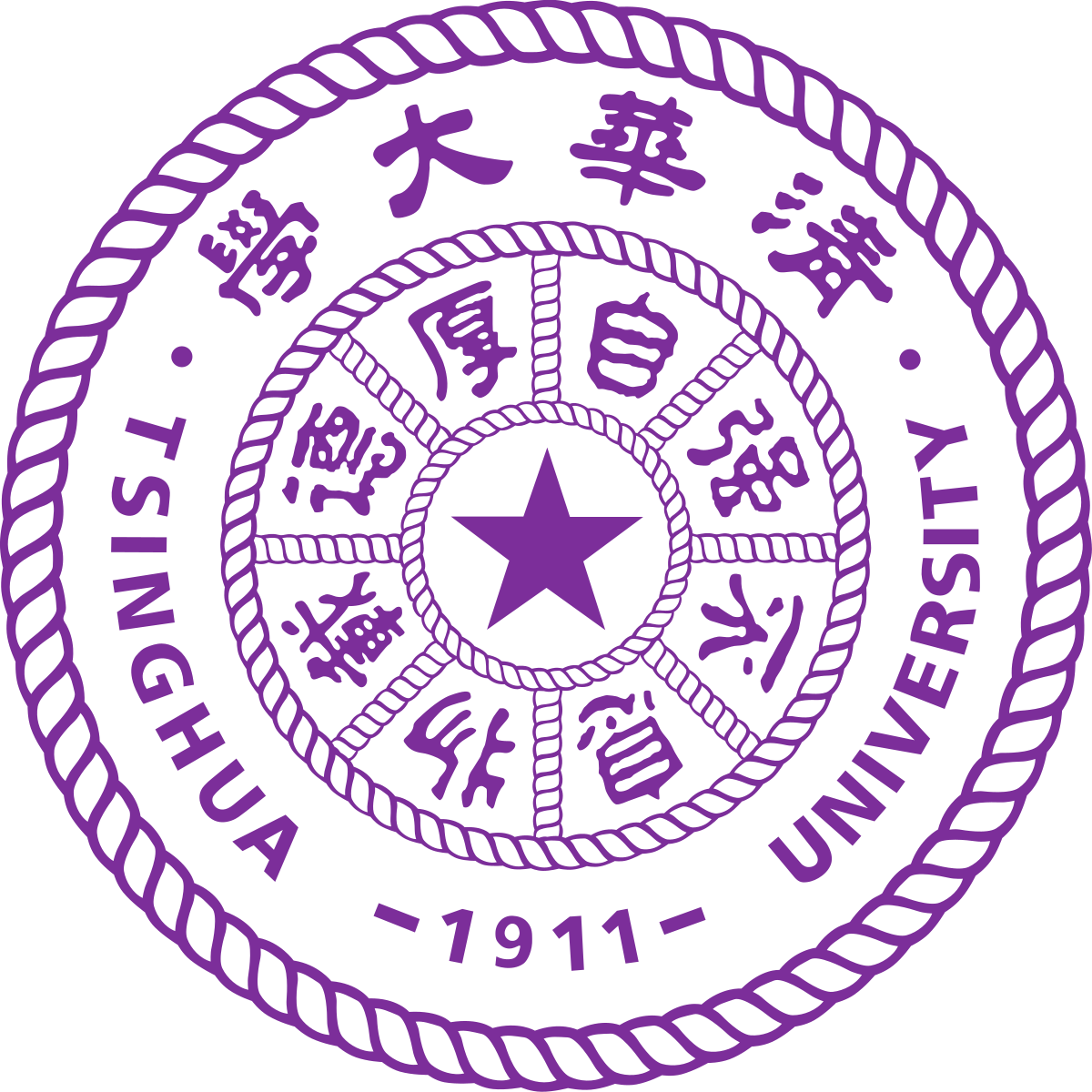}}
\newcommand{\squad}{\hspace{0.4em}}

\usepackage{algorithm,algorithmicx,algpseudocode}

\newcolumntype{C}{>{$\displaystyle}c<{$}}

\definecolor{citecolor}{HTML}{0b64c5}

\definecolor{cello}{HTML}{f2ffea}
\newcommand{\colorcello}{\cellcolor{cello}}

\definecolor{lightblue}{HTML}{e1f4ff}

\newcommand{\cellcolorlightblue}{\cellcolor{lightblue}}

\definecolor{lightgray}{HTML}{f2f2f2}
\newcommand{\colorlightgray}{\colorbox{lightgray}}
\newcommand{\cellcolorlightgray}{\cellcolor{lightgray}}

\definecolor{lightpurple}{HTML}{f2eaff}
\newcommand{\colorlightpurple}{\colorbox{lightpurple}}
\newcommand{\cellcolorlightpurple}{\cellcolor{lightpurple}}

\definecolor{lightgreen}{HTML}{f2ffea}
\newcommand{\colorlightgreen}{\colorbox{lightgreen}}
\newcommand{\cellcolorlightgreen}{\cellcolor{lightgreen}}

\definecolor{sp_lightgreen}{HTML}{e2f0d9}

\definecolor{sp_lightorange}{HTML}{fbe5d6}

\definecolor{lblue}{rgb}{0.2, 0.5, 0.9}

\def\paperID{10391} 
\def\confName{CVPR}
\def\confYear{2024}

\DeclareUnicodeCharacter{1EAF}{\'a}

\title{\includegraphics[scale=0.007]{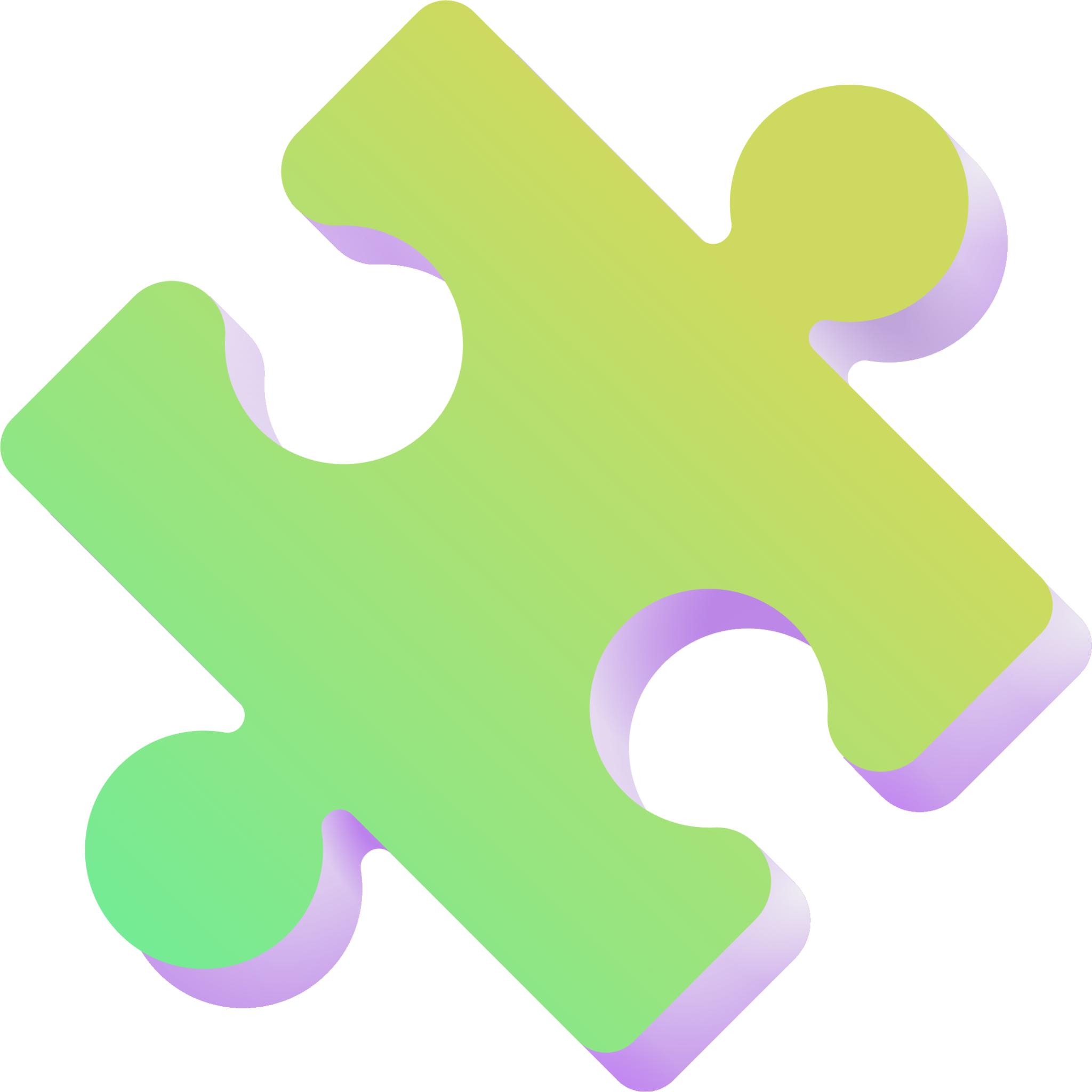} \textsc{\gradientRGB{TokenCompose}{201,219,97}{6,60,255}}: Text-to-Image Diffusion with Token-level Supervision}
\author{%
  Zirui Wang$^{\text{\puicon},\text{\ucsdicon}}$ \squad Zhizhou Sha$^{\text{\tsinghuaicon},\text{\ucsdicon}}$ \squad Zheng Ding$^{\text{\ucsdicon}}$ \squad
  Yilin Wang$^{\text{\tsinghuaicon},\text{\ucsdicon}}$ \squad Zhuowen Tu$^{\text{\ucsdicon}}$ \\
  $\text{\puicon}$ Princeton University \squad $\text{\tsinghuaicon}$ Tsinghua University \squad $\text{\ucsdicon}$ University of California, San Diego \\
  \url{https://mlpc-ucsd.github.io/TokenCompose}
}

\begin{document}
\twocolumn[{
\renewcommand\twocolumn[1][]{#1}
\maketitle
\vspace*{-1.2cm}
\begin{center}
    \centering
    \includegraphics[width=\textwidth]{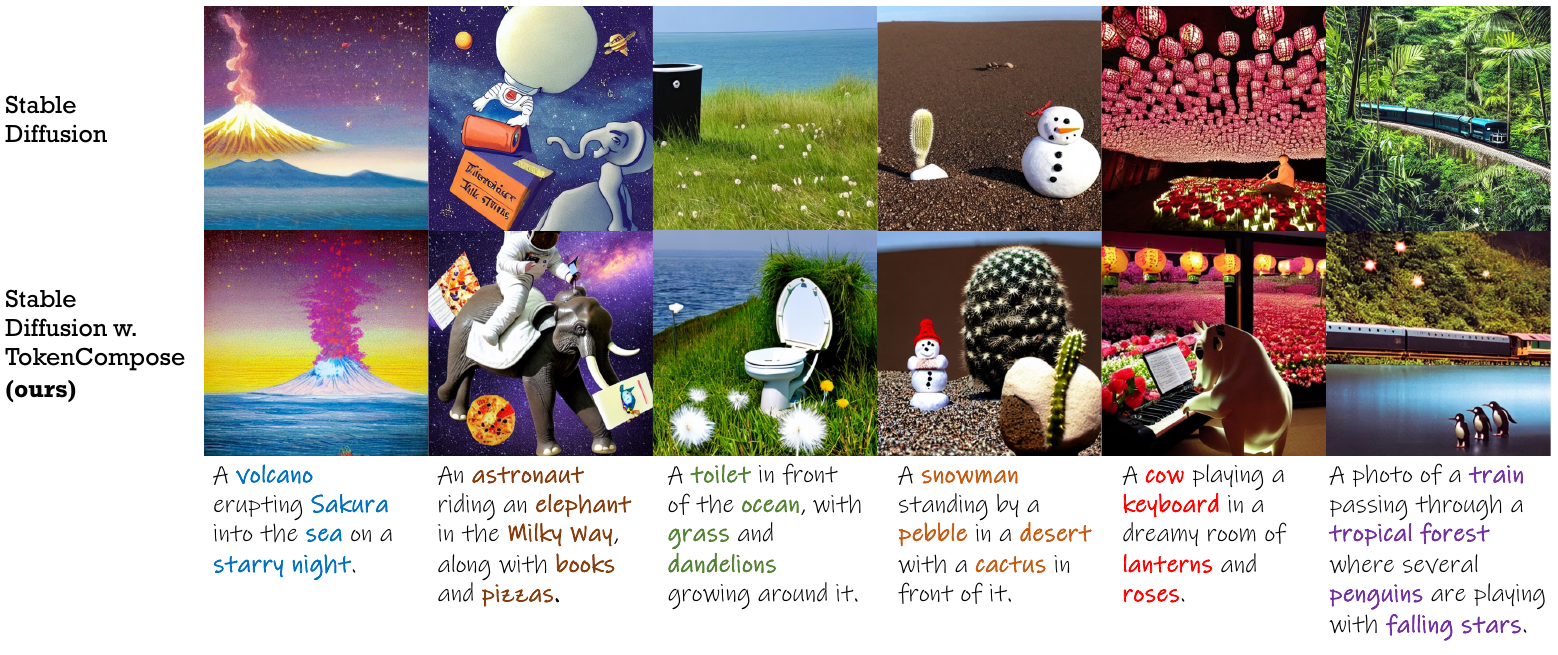}
    \vspace*{-1cm}
    \captionsetup[figure]{hypcap=false}
    \captionof{figure}{\textbf{Given a user-specified text prompt consisting of object compositions} that are \emph{unlikely} to appear simultaneously in a natural scene, our proposed \textsc{TokenCompose} method attains significant performance enhancement over the baseline Latent Diffusion Model (\textit{e.g.,} Stable Diffusion~\cite{ldm}) by being able to generate multiple categories of instances from the prompt more accurately.}
\label{fig:teaser}
\end{center}
\vspace*{-1.6ex}
}]

\newcommand{\projectpath}{https://github.com/CompVis/latent-diffusion}

\newcommand*{\approxident}{%
  \mathrel{\vcenter{\offinterlineskip
  \hbox{$\sim$}\vskip-.35ex\hbox{$\sim$}\vskip-.35ex\hbox{$\sim$}}}}
\newcommand{\RR}{\mathbb{R}}
\newcommand{\expec}{\mathbb{E}}
\newcommand{\prior}{\mathcal{N}(0,1)}

\newcommand{\xpixel}{x_\text{pixel}}
\newcommand{\xpixelrec}{\tilde{x}_\text{pixel}}
\newcommand{\hpixel}{H}
\newcommand{\wpixel}{W}
\newcommand{\cpixel}{3}
\newcommand{\xrec}{\tilde{x}}

\newcommand{\latent}{z_0}
\newcommand{\hlatent}{h}
\newcommand{\wlatent}{w}
\newcommand{\clatent}{c}

\newcommand{\zt}[1]{z_{#1}}
\newcommand{\enc}{E}
\newcommand{\ppixel}{p_\text{pixel}}
\newcommand{\pzt}[1]{p_{\zt{#1}}}
\newcommand{\qzt}[1]{q_{\zt{#1}}}
\newcommand{\q}{q}
\newcommand{\R}{\mathbb{R}}

\newcommand{\LPIPS}{\text{LPIPS}}
\newcommand{\KL}{\mathbb{KL}}
\newcommand{\expect}{\mathbb{E}}
\newcommand{\pmodel}[1]{p^{#1}_{\theta}}
\newcommand{\pchain}{p_{\theta}}
\newcommand{\qchain}{q}
\newcommand{\qmodel}[1]{q_{#1}}

\newcommand{\qenc}{q_{\phi}}
\newcommand{\pdec}{p_{\phi}}
\newcommand{\dec}{G_{\phi}}

\newcommand{\disc}{D_{\psi}}

\newcommand{\lrec}{L_{rec}}
\newcommand{\ladv}{L_{adv}}
\newcommand{\lreg}{L_{reg}}
\newcommand{\lcomp}{L_{cm}}
\newcommand{\lsimple}{\mathcal{L}_{DM}}
\newcommand{\lsimpleldm}{\mathcal{L}_{LDM}}
\newcommand{\lsimplelcm}{\mathcal{L}_{LDM}}

\newcommand{\model}{\epsilon_\theta}
\newcommand{\conditioner}{\tau_\theta}

\newcommand{\encoder}{\mathcal{E}}
\newcommand{\decoder}{\mathcal{D}}

\newcommand{\cond}{y}

\begin{abstract}
\vspace{-3ex}

We present TokenCompose, a Latent Diffusion Model for text-to-image generation that achieves enhanced consistency between user-specified text prompts and model-generated images. Despite its tremendous success, the standard denoising process in the Latent Diffusion Model takes text prompts as conditions only, absent explicit constraint for the consistency between the text prompts and the image contents, leading to unsatisfactory results for composing multiple object categories. Our proposed TokenCompose aims to improve multi-category instance composition by introducing the token-wise consistency terms between the image content and object segmentation maps in the finetuning stage. TokenCompose can be applied directly to the existing training pipeline of text-conditioned diffusion models without extra human labeling information. By finetuning Stable Diffusion with our approach, the model exhibits significant improvements in multi-category instance composition and enhanced photorealism for its generated images. \footnote{Project done while ZW, ZS and YW interned at UC, San Diego. Correspondence to ZW at \href{mailto:zw1300@cs.princeton.edu}{\texttt{zw1300@cs.princeton.edu}}}

\end{abstract}

\section{Introduction}
\label{sec:intro}
\vspace{-1.2ex}

Despite the tremendous progress in recent text-to-image diffusion models \cite{dalle,ldm, controlnet,ruiz2023dreambooth, imagen, feng2023ernie, xue2024raphael, chen2024pixartalpha, hertz2022prompt} that have achieved creating images with an increasing level of quality, resolution, photorealism, and diversity, there still exists a major consistency problem between the text prompt and the generated image content. The models loose the composition capability when multiple object categories, especially those not commonly appearing simultaneously in the real world, are included in the text prompt: objects may not appear in the image or their configuration is not pleasantly good looking. Figure \ref{fig:teaser} shows examples where a state-of-the-art model, Stable Diffusion \cite{ldm}, fails to generate desirable image content from text prompts.

\begin{figure*}[ht!]
  \centering
    \includegraphics[width=1\linewidth]{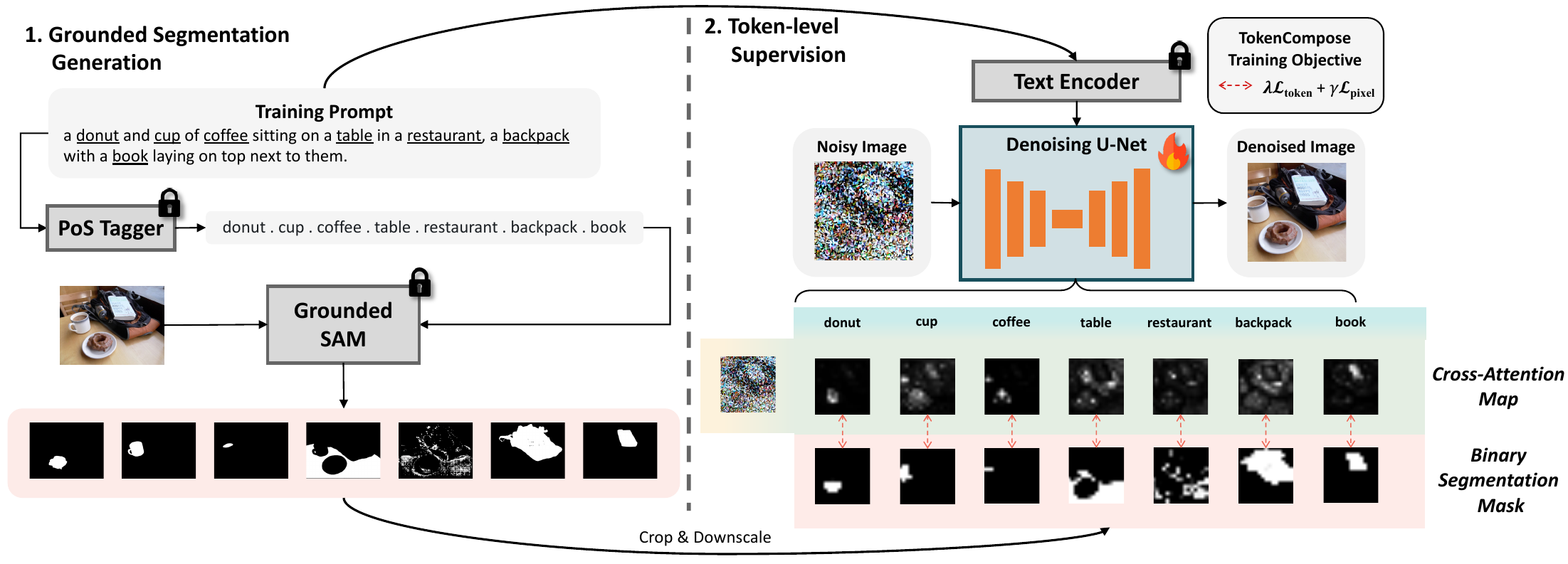}
    \vspace*{-4ex}
    \caption{\textbf{An overview of the} \textsc{TokenCompose} \textbf{training pipeline.} Given a training prompt that faithfully describes an image, we adopt a POS tagger \cite{flair} and Grounded SAM \cite{sam, grounding_dino} to extract all binary segmentation maps of the image corresponding to noun tokens from the prompt. Then, we jointly optimize the denoising U-Net of the diffusion model with both its original denoising and our grounding objective.}
    \label{fig:method}
\vspace*{-4ex}
\end{figure*}

Prototypical generative models of various families \cite{vae, gan, tu2007learning, introspective, norm_flow, ddpm, ldm, bayesian} have reached maturity. Adding conditional training signals \cite{glide, gligen, cogview, cogview2, dalle, dalle_mini, ldm, controlnet, karras2019style} to the generative models significantly expands their modeling capability, as well as their scope of application. In the context of Latent Diffusion Models \cite{ldm}, one of the most commonly applied conditions, text (\textit{e.g.,} captions), is injected into layers of the denoising U-Net \cite{unet} via cross-attention. 

However, there exists a natural discrepancy between texts (\textit{e.g.,} captions) that are used to train such models and texts (\textit{e.g.,} prompts) that are used for generation. Whereas a caption usually describes a real image faithfully, a prompt can encapsulate image features that do not match the visual scene of any real-world images. Without fine-grained training objectives for its conditioned text, a text-to-image diffusion model often fails to generalize to arbitrary compositions that lie in the prompts \cite{mm_failures}. This can be due to the fact that the denoising objective in text-to-image LDM is only optimized to predict the noise given a text prompt, leaving the text condition only as a facilitation in optimizing the denoising function.


By imposing training objectives that operate at the token level in the conditional text, the diffusion model learns what each token from the text means in the context of the image in an \textit{atomistic} manner. Subsequently, it can be better at composing different combinations of words, phrases, \etc, during inference. However, obtaining the ``ground-truth'' labels (\textit{e.g.,} segmentation maps) by humans for the corresponding text tokens is label intensive and expensive, especially for the text-image pairs \cite{laion} used to train large-scale generative models. Thanks to the recent progress in vision foundation models, Grounding DINO \cite{grounding_dino} and Segment Anything (SAM) \cite{sam}, grounding segmentation maps for text tokens can be readily attained automatically.

To this end, we seek to mitigate the composition problem by developing a new algorithm, \textsc{TokenCompose}, which leverages models pretrained with image understanding tasks \cite{sam, grounding_dino, flair} to provide token-level training supervisions to a text-to-image generative model. We show that, by augmenting each noun token from the text prompt of a text-to-image model with segmentation grounding objectives with respect to its respective image during training, the model exhibits significant improvement in object accuracy \cite{visor}, multi-category instance composition, enhanced photorealism \cite{fid} with no additional inference cost for its generated images. Along with our proposed training framework for text-to-image generative models, we also present the \textsc{MultiGen} benchmark, which examines the capabilities of a text-to-image generative model to compose multiple categories of instances in a single image. 

\vspace{-0.8ex}
\section{Related Works}
\label{sec:related_works}
\vspace{-0.8ex}

\textbf{Compositional Generation.} Efforts that aim to improve compositional image generation for text-conditioned image generative models have been focused on both the training and inference stages. One approach to improving the compositional generation of diffusion models in training is by introducing additional modules, such as a ControlNet, to specify high-level features within the image \cite{controlnet, control_gpt}. However, the modules added to the diffusion models increase the size of the model, leading to additional training and inference costs. Another approach through training is to leverage a reward function to encourage faithful generation of images based on compositional prompts \cite{t2i_bench, imagereward, t2i_hf, diffusion_rl}. Albeit their efficacy, reward functions are sparse and do not provide dense supervision signals. 

Inference-based methods aim to alter the latent and/or cross-attention maps. Composable Diffusion \cite{composable_diffusion} decomposes a compositional prompt into sub-prompts to generate different latents, and uses a score function to combine the latents together, while Layout Guidance Diffusion \cite{layout_diffusion} uses user-defined tokens and bounding boxes to backpropagate gradients to the latent, and steer the cross-attention maps to focus on specified regions for specified tokens. Other methods apply Gaussian kernels \cite{attend_and_excite} or leverage linguistic features \cite{structured_diffusion, linguistic_binding} to manipulate the cross-attention map. While these methods do not require further training, they add considerable cost during inference, making it cost $\times$3.37 times longer to generate a single image at most, when other generation configurations are kept the same.

Our framework is training-based and does not require additional modules to be incorporated into the image generation pipeline. Furthermore, optimizing attention maps based on segmentation maps provides dense and interpretable supervision. As a training-based method that jointly optimizes token-image correspondence and image generation, the model does not require inference-time manipulations, yet it achieves strong compositionality and competitive image quality for conditional generation.

\textbf{Benchmarks for Compositional Generation. } Several benchmarks have been proposed to evaluate text-conditioned image generative models for compositionality. Most methods evaluate compositionality by binding attributes to and specifying relations between objects. For example, spatial relationships examine whether two different objects appear in the correct spatial layout according to the prompt \cite{visor, hrs_bench, t2i_bench, paintskills, imagen}. Color binding examines whether a text-conditioned image generative model can correctly assign the specified colors to different objects, especially when color assignments are \emph{counterintuitive} \cite{structured_diffusion, attend_and_excite, hrs_bench, t2i_bench, imagen}. Count binding examines whether a specified instance appears with the right number of counts specified in the prompt \cite{imagen, hrs_bench, paintskills}. There are additional types of attribute bindings and relation specifications that are evaluated in different benchmarks, such as action and size \cite{hrs_bench}, and shape and texture \cite{t2i_bench}.

However, the majority of such benchmarks confound evaluating capabilities of binding the correct attributes or specifying the correct inter-object relationship with the successful generation of specified objects mentioned in the prompt, making it difficult to evaluate whether improvements are made by stronger attribute assignment \& relation specification capabilities or a higher object accuracy \cite{visor, paintskills, soa}. VISOR \cite{visor} decouples object accuracy from spatial relationship compositionality by calculating the successful rate of correct spatial relationships conditioned on the successful generation of all specified instances from the prompt. In contrast, almost all other benchmarks do not dissociate these factors and instead evaluate compositionality on a \emph{holistic} basis.

Existing benchmarks on multi-category instance composition focus on successful generation of mostly two categories. On the other hand, leading image generative models have achieved significant improvements in multi-category instance composition \cite{dalle3, recaption}, which are capable of generating multiple categories of instances with a high success rate. To fill in this research gap in evaluating multi-category instance composition beyond two categories, and to evaluate our training framework in multi-category instance composition, we propose \textsc{MultiGen} benchmark, which contains text prompts where each prompt contains objects from an arbitrary combination of multiple categories.

\textbf{Generative Models for Image Understanding. } There has been an upward trend in the use of text-conditioned image generative models for open-vocabulary image understanding tasks, such as classification \cite{secret_classifier, diffusion_zs_classification}, detection \cite{ovtrack, diffusiondet}, and segmentation \cite{ov_obj_seg_diffusion, odise, diffuse_attend_segment, slime, diffusionseg, ref_diff, secret_segmenter, text_to_mask, daam, wu2023diffumask, zhao2023unleashing}.

An inherent advantage of using generative models for image understanding tasks is being open-vocabulary, as text-to-image diffusion models are trained with open-vocabulary textual prompts. Although results using Stable Diffusion \cite{ldm} for unsupervised zero-shot image understanding show great potential among methods in the same setting (\textit{i.e.,} zero-shot and open-vocabulary), there still exists a performance gap between these approaches and those that use specialist models for image understanding \cite{coca, dino, grounding_dino, sam, mask2former}. This gives us the empirical basis to optimize a generative model (\textit{e.g.,} Stable Diffusion) with knowledge from an understanding model. Furthermore, by optimizing a diffusion model with our approach, we also observe improved performance for downstream segmentation tasks \cite{daam, coco}, which further reflects successful transfer of this knowledge.

\vspace{-1ex}
\section{Method}
\vspace{-0.8ex}

\subsection{Preliminaries}
Diffusion models \cite{ddpm} are widely used in conditional image generative tasks. Given an image $x \in \mathbb{R}^{\hpixel \times \wpixel \times \cpixel}$, a normally distributed variable $\epsilon$ (\textit{e.g.,} noise) is added to the image with a variable extent based on a timestep $t$. Given a denoising function parameterized by a neural network $\theta$, a noisy image $x_t$, and a timestep $t$ uniformly sampled from $\{1, \dots, T\}$, the denoising function learns to predict the noise, $\epsilon$, following the objective (Eq. \ref{eq:dm}):

\vspace{-3mm}
\begin{equation}
\label{eq:dm}
\lsimple = \expec_{x, \epsilon \sim \mathcal{N}(0, 1),  t }\Big[ \Vert \epsilon - \model(x_{t},t) \Vert_{2}^{2}\Big] \,
\vspace{-1mm}
\end{equation}

To improve the efficiency and control of diffusion models, two changes are performed on the original diffusion recipe -- forming the Latent Diffusion Model \cite{ldm}. 

First, instead of learning a denoising function in the image space, an image is encoded in a latent state $z_0 = \encoder(x_0)$ using a variational autoencoder (VAE) \cite{vae}. A random noise $\epsilon$ is added to the latent $z_0$, resulting in a noisy latent $z_t$. The training process involves computing the loss between the predicted noise $\epsilon_\theta$ and the ground truth noise $\epsilon$ to optimize the denoising function.

Second, a conditioning mechanism is added to the denoising function to steer the diffusion process for controllable image generation via cross-attention. In our setting, the condition $y$ is a text prompt that describes the image. To use the text via cross-attention, each token is transformed into an embedding $\conditioner(y)$ using a pretrained text encoder \cite{clip, openclip}. The following shows the denoising objective for an LDM (Eq. \ref{eq:ldm}):

\vspace{-6mm}
\begin{equation}
\lsimplelcm := \expec_{\encoder(x), y, \epsilon \sim \mathcal{N}(0, 1), t }\Big[ \Vert \epsilon - \model(z_{t},t, \conditioner(y)) \Vert_{2}^{2}\Big] \,
\label{eq:ldm}
\vspace{-2mm}
\end{equation}

We only optimize the denoising function $\model$, which is parameterized by a U-Net \cite{unet} architecture during training. $\encoder$, $\decoder$, and $\conditioner$ are kept frozen.

\subsection{Token-level Attention Loss}

Consider a text prompt that is transformed into text embeddings of length $L_{\conditioner(y)}$. As $\lsimplelcm$ only optimizes the function so that it predicts the noise and reconstructs the image latent by removing the noise, the relationship between each token's embedding $e_i$, $i \in \{1, \dots, L_{\conditioner(y)}\}$ and a noisy image latent $z_t$ is not optimized explicitly. This leads to a poor token-level understanding in an LDM, which can be visualized via activations of the multihead cross-attention map (\textit{i.e.,} $\mathcal{A}$) between the token's embedding (\textit{i.e.,} $K \in \mathbb{R}^{H \times L_{\conditioner(y)} \times d_k}$), and the noisy image latent (\textit{i.e.,} $Q \in \mathbb{R}^{H \times L_{z_{t}} \times d_k}$). For each cross-attention layer $m \in M$ with variable latent representation resolutions $L_{z_{t}}$ in the U-Net, the cross-attention map (\textit{i.e.,} $\mathcal{A} \in \mathbb{R}^{L_{z_{t}} \times L_{\conditioner(y)}}$) is calculated as the following (Eq. \ref{eq:qk_calc} and \ref{eq:avg_then_loss}):

\vspace{-6mm}
\begin{equation}
Q^{(h)} = W^{(h)}_Q \cdot  \varphi(z_t), \; K^{(h)} = W^{(h)}_K \cdot \conditioner(y)\;
\label{eq:qk_calc}
\vspace{-2mm}
\end{equation}

\vspace{-3mm}
\begin{equation}
\mathcal{A} = \frac{1}{H} \sum_{h}^{H} \text{softmax}\left (\frac{Q^{(h)}(K^{(h)})^T}{\sqrt{d_k}}\right )
\label{eq:avg_then_loss}
\vspace{-2mm}
\end{equation}

where $h \in \{1, \dots, H\}$ represents each head in the multihead cross-attention, $\varphi$ is a function that flattens a two-dimensional image latent into one dimension, and $d_k$ is the dimension of $K$.

\begin{figure}[ht!]
  \centering
    \includegraphics[width=1\linewidth]{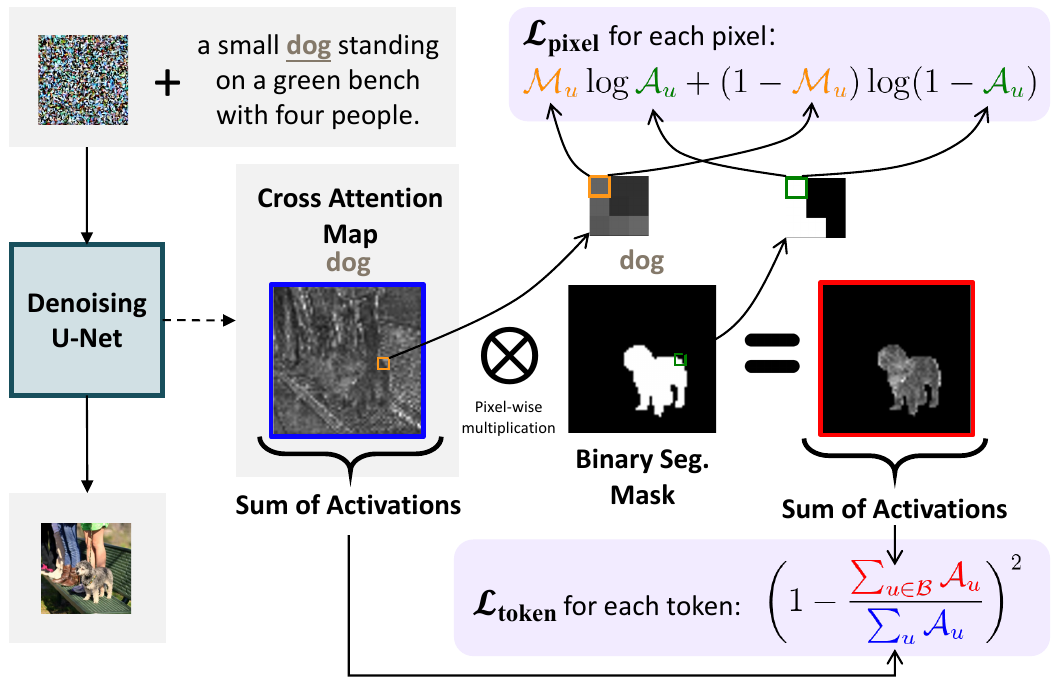}
    \vspace{-4ex}
    \caption{\textbf{Illustration of $\mathcal{L}_\text{token}$ and $\mathcal{L}_\text{pixel}$.} We illustrate how $\mathcal{L}_\text{token}$ and $\mathcal{L}_\text{pixel}$ are calculated given a cross-attention map $\mathcal{A}_i$ and a binary segmentation mask $\mathcal{M}_i$. $\mathcal{L}_\text{token}$ aggregates attention activations toward non-masked regions of $\mathcal{M}_i$, and this objective is normalized by the total activations of $\mathcal{A}_i$. However, it does \emph{not} constrain where activations should be once inside the non-masked region. $\mathcal{L}_\text{pixel}$ gives precise supervision whether a pixel belongs to the segmented region, constraining where activations should be with binary values. However, it is \emph{not} normalized by the total activations of $\mathcal{A}_i$. Combining $\mathcal{L}_\text{token}$ and $\mathcal{L}_\text{pixel}$, we take advantage of the benefit of each objective while minimizing their side effects to a minimum level. We show examples of cross-attention activations from models optimized with \colorlightgreen{$\mathcal{L}_\text{token}$ and $\mathcal{L}_\text{pixel}$}, \colorlightpurple{either} of them, and \colorlightgray{neither} of them in Figure \ref{fig:activation_viz}.}
    \label{fig:failure_demonstration}
\vspace{-4ex}
\end{figure}

Empirically, we observe that training a diffusion model with only $\lsimplelcm$ often causes activations of cross attention maps of distinct instance tokens to fail to focus on its corresponding instance appeared in the image during training, which, in turn, results in poor capabilities in composing multiple categories of instances during inference.

To alleviate this issue for better multi-category instance composition, we add a training constraint that supervises activation regions of the cross-attention maps. Specifically, for each text token $i$ that belongs to a noun within the text caption, we acquire the binary segmentation map $\mathcal{M}_i$ from its respective image by leveraging foundation models trained for image understanding \cite{grounding_dino, sam}. Because cross-attention maps at different layers $m$ of the U-Net have different resolutions, we downscale the resolution of $\mathcal{M}_i$ to match the dimensions of its corresponding $\mathcal{A}_i^{(m)}$ with bilinear interpolation, followed by binarization of all values to form $\mathcal{M}_i^{(m)}$. Different from Layout Diffusion \cite{layout_diffusion}, which uses user-defined bounding boxes during inference for gradient-based guidance, we directly apply a loss function $\mathcal{L}_\text{token}$ that aggregates activations of cross-attention toward predicted spatial regions $\mathcal{B}_i = \{u \in \mathcal{M}_i \mid u = 1 \}$ jointly with $\lsimplelcm$ during model training. For any layer $m$, $\mathcal{L_\text{token}}$ is defined as follows (Eq. \ref{eq:token_loss}):

\vspace{-3ex}
\begin{equation}
\label{eq:token_loss}
\mathcal{L}_\text{token} = \frac{1}{N}\sum_{i}^{N} \left (1 - \frac{\sum_{u \in \mathcal{B}_i}^{L_{z_{t}}}\mathcal{A}_{(i, u)}}{\sum_{u}^{L_{z_{t}}}\mathcal{A}_{(i, u)}} \right )^{2}
\end{equation}

 where $\mathcal{A}_{(i, u)} \in \mathbb{R}$ represents the scalar attention activation at a spatial location $u$ of $\mathcal{A}_i \in \mathbb{R}^{L_{z_t}}$ for the cross-attention map formed by the latent and the $i$th token's embedding. Whereas the previous approach \cite{layout_diffusion} calculates the loss on each attention head of a multihead cross-attention module separately, we calculate the loss on the average of cross-attention activations on all heads (see Eq. \ref{eq:avg_then_loss}). We find that the latter approach encourages different heads to activate in distinct regions of the cross-attention map, which slightly improves compositional performance and image quality. We add a scaling factor $\lambda$ to $\mathcal{L_\text{token}}$ with respect to $\lsimplelcm$ such that sufficient token-level gradients can be used to optimize token-image consistency while the denoising objective is minimally compromised.

\subsection{Pixel-level Attention Loss}
\begin{figure}[ht!]
  \centering
    \includegraphics[width=1\linewidth]{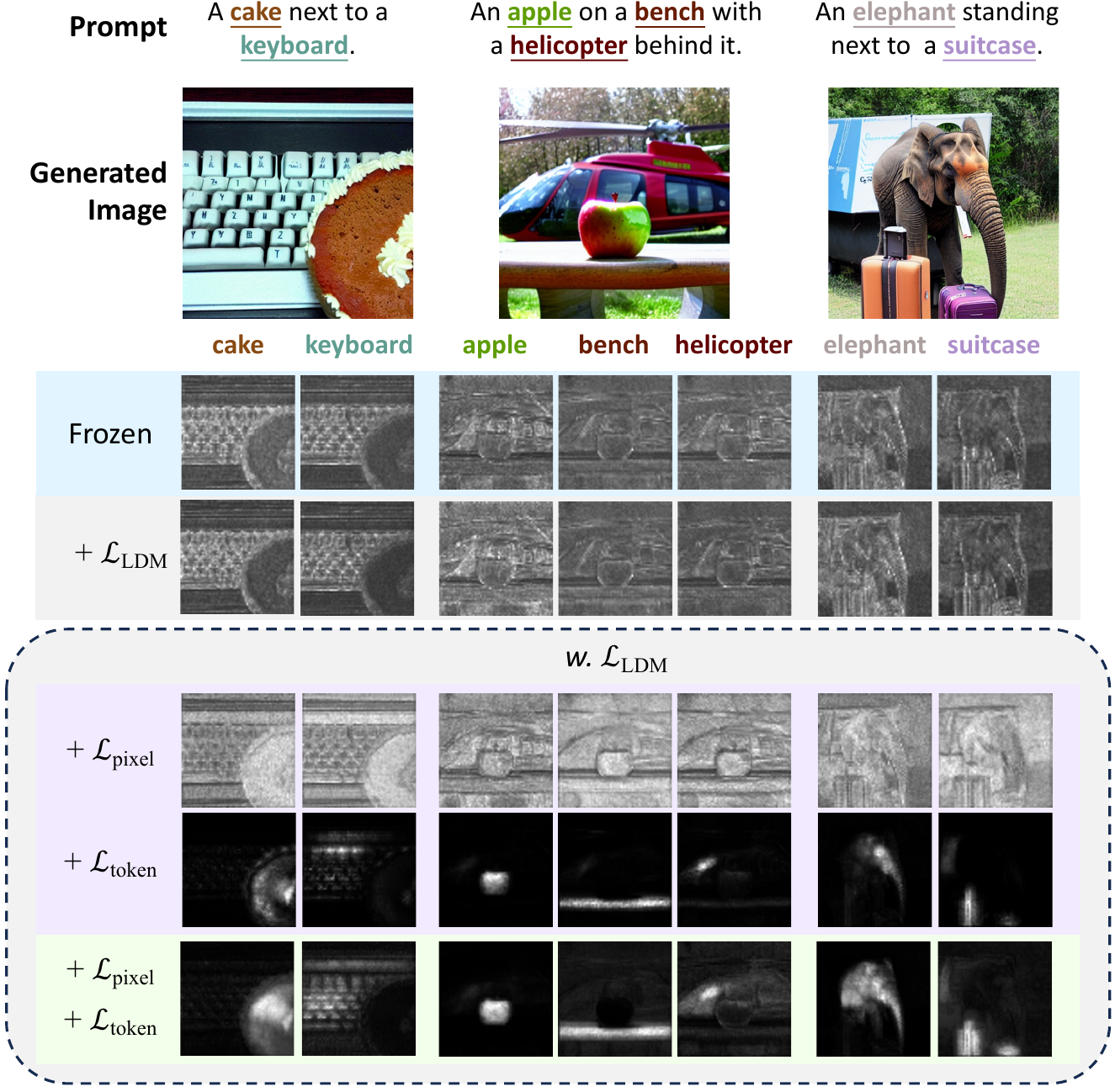}
    \vspace{-4ex}
    \caption{\textbf{Impact on cross-attention activations with different objectives.} We firstly demonstrate that finetuning the Stable Diffusion with only \colorlightgray{$\lsimplelcm$} does not improve grounding capabilities as much. Adding \colorlightpurple{$\mathcal{L_\text{pixel}}$} alone causes increased cross-attention activations in general. Adding \colorlightpurple{$\mathcal{L_\text{token}}$} plays a vital role in improving token grounding, but leads activations to aggregate in subregions of the targets. By combining \colorlightgreen{$\mathcal{L_\text{token}}$ and $\mathcal{L_\text{pixel}}$}, the model shows substantial improvement in grounding text tokens with image features. In this illustration, we apply the null text inversion \cite{null_text_inversion} technique to all models, allowing them to generate the same image for comparable cross-attention maps.}
    \label{fig:activation_viz}
\vspace{-2ex}
\end{figure}

Although $\mathcal{L_\text{token}}$ substantially aggregates activations of cross-attention maps toward the target regions, a side effect of this aggregation is that the model tends to overly aggregate its activations of the cross-attention map into certain subregions of its target regions. This can be reflected by visually inspecting its cross-attention map during inference (see Figure \ref{fig:activation_viz}) and an increased binary cross-entropy loss of cross-attention map activations $\mathcal{A}$ with respect to the target binary segmentation map $\mathcal{M}$. To overcome this problem, we use $\mathcal{L}_{\text{pixel}}$ to counteract. Formally, for a cross-attention map $\mathcal{A}$ in any layer $m$ optimized with $\mathcal{L_\text{token}}$, we add the pixel-level cross-entropy objective that is defined as the following (Eq. \ref{eq:bce}):

\vspace{-6ex}
\begin{align}
\label{eq:bce}
\mathcal{L}_\text{pixel} &= -\frac{1}{L_{\conditioner(y)}L_{z_t}}\sum_{i}^{L_{\conditioner(y)}}\sum_{u}^{L_{z_t}} \left (\mathcal{M}_{(i, u)} \log(\mathcal{A}_{(i, u)} \right ) \\ \notag &\quad + \left (1 - \mathcal{M}_{(i, u)} \right ) \log \left (1-\mathcal{A}_{(i, u)} \right )
\end{align}

\vspace{-2ex}

We add a scaling factor $\gamma$ so that $\mathcal{L}_{\text{pixel}}$ is kept roughly constant while the model is jointly optimized by $\lsimplelcm$ and $\mathcal{L_\text{token}}$. We provide a token- and pixel-level optimization illustration in Figure \ref{fig:failure_demonstration} to demonstrate different levels of granularity of $\mathcal{L_\text{token}}$ and $\mathcal{L_\text{pixel}}$. Finally, at any single optimization step, the training objective is as follows (Eq. \ref{eq:tc_loss}):

\vspace{-5mm}
\begin{equation}
\label{eq:tc_loss}
\mathcal{L}_\textsc{TokenCompose} = \underbrace{\lsimplelcm}_\text{denoise} + \sum_{m}^{M} \left (\underbrace{\lambda\mathcal{L}_\text{token}^{(m)}}_\text{token grouding} + \underbrace{\gamma\mathcal{L}_\text{pixel}^{(m)}}_{\text{pixel grounding}}\right )
\end{equation}
\vspace{-9mm}

\vspace{-2.72ex}
\section{Experiment}

\begin{table*}
  \centering
  \scalebox{0.73}{
  \begin{tabular}{llc|cccc|cccccccccc@{}}
    \hline
    \toprule
    \multicolumn{1}{c}{\textbf{}} & \phantom{} & \multicolumn{9}{c}{\textbf{Multi-category Instance Composition $(\uparrow)$}} & \phantom{} &  \multicolumn{2}{c}{\textbf{Photorealism $(\downarrow)$}} & \phantom{} & \multicolumn{1}{c}{\textbf{Eff. $(\downarrow)$}} \\

    \cmidrule{3-11} \cmidrule{13-14} \cmidrule{16-16} 
    
    \textbf{Method} & \phantom{} & Object & \multicolumn{4}{c|}{\textsc{COCO Instances}}  & \multicolumn{4}{c}{\textsc{ADE20K Instances}} & \phantom{} & FID & FID & \phantom{} &Latency \\
    
     & \phantom{} & Accuracy & MG2 & MG3 & MG4 & MG5 & MG2 & MG3 & MG4 & MG5 & \phantom{} & (\textbf{C}) & (\textbf{F}) & \phantom{} & \\
    \midrule
    
    SD \cite{ldm} & \phantom{} & 29.86 & 90.72\textsubscript{1.33} & 50.74\textsubscript{0.89} & 11.68\textsubscript{0.45} & 0.88\textsubscript{0.21} & 89.81\textsubscript{0.40} & 53.96\textsubscript{1.14} & 16.52\textsubscript{1.13} & 1.89\textsubscript{0.34} & \phantom{} & \uline{20.88} & \uline{71.46} & \phantom{} & \textbf{7.54}\textsubscript{0.17}\\
    
    Composable \cite{composable_diffusion} & \phantom{} & 27.83 & 63.33\textsubscript{0.59} & 21.87\textsubscript{1.01} & 3.25\textsubscript{0.45} & 0.23\textsubscript{0.18} & 69.61\textsubscript{0.99} & 29.96\textsubscript{0.84} & 6.89\textsubscript{0.38} & 0.73\textsubscript{0.22} & \phantom{} & - & 75.57 & \phantom{} & 13.81\textsubscript{0.15}\\
    
    Layout \cite{layout_diffusion} & \phantom{} & 43.59 & 93.22\textsubscript{0.69} & 60.15\textsubscript{1.58} & 19.49\textsubscript{0.88} & 2.27\textsubscript{0.44} & \uline{96.05}\textsubscript{0.34} & \uline{67.83}\textsubscript{0.90} & 21.93\textsubscript{1.34} & 2.35\textsubscript{0.41} & \phantom{} & - & 74.00 & \phantom{} & 18.89\textsubscript{0.20}\\
    
    Structured \cite{structured_diffusion} & \phantom{} & 29.64 & 90.40\textsubscript{1.06} & 48.64\textsubscript{1.32} & 10.71\textsubscript{0.92} & 0.68\textsubscript{0.25} & 89.25\textsubscript{0.72} & 53.05\textsubscript{1.20} & 15.76\textsubscript{0.86} & 1.74\textsubscript{0.49} & \phantom{} & 21.13 & 71.68 & \phantom{} & \uline{7.74}\textsubscript{0.17} \\
    
    Attn-Exct \cite{attend_and_excite} & \phantom{} & \uline{45.13} & \uline{93.64}\textsubscript{0.76} & \uline{65.10}\textsubscript{1.24} & \uline{28.01}\textsubscript{0.90} & \textbf{6.01}\textsubscript{0.61} & 91.74\textsubscript{0.49} & 62.51\textsubscript{0.94} & \uline{26.12}\textsubscript{0.78} & \uline{5.89}\textsubscript{0.40} & \phantom{} & - & 71.68 & \phantom{} & 25.43\textsubscript{4.89} \\
    
    \midrule
    \colorcello \emph{\textbf{Ours}} & \colorcello \phantom{} & \colorcello \textbf{52.15} & \colorcello \textbf{98.08}\textsubscript{0.40} & \colorcello \textbf{76.16}\textsubscript{1.04} & \colorcello \textbf{28.81}\textsubscript{0.95} & \colorcello \uline{3.28}\textsubscript{0.48} & \colorcello \textbf{97.75}\textsubscript{0.34} & \colorcello \textbf{76.93}\textsubscript{1.09} & \colorcello \textbf{33.92}\textsubscript{1.47} & \colorcello \textbf{6.21}\textsubscript{0.62} & \colorcello \phantom{} & \colorcello \textbf{20.19} & \colorcello \textbf{71.13} & \colorcello \phantom{} & \colorcello \textbf{7.56}\textsubscript{0.14}\\
    \bottomrule
    \hline
  \end{tabular}
  }
  \vspace{-1ex}
  \caption{\textbf{Performance of our model in comparison to baselines.} We evaluate the performance based on multi-category instance composition (\textit{i.e.,} Object Accuracy (OA) from VISOR Benchmark \cite{visor} and MG2-5 from our \textsc{MultiGen} Benchmark), photorealism (\textit{i.e.,} FID \cite{fid} from \textbf{C}OCO and \textbf{F}lickr30K Entities validation splits), and inference efficiency. All comparisons are based on Stable Diffusion 1.4.}
  \label{tab:main_result}
\vspace{-4ex}
\end{table*}

\subsection{Training Details} 
\hypertarget{Section4.1_Dataset}{\textbf{Dataset.}} To study the effectiveness of token grounding objectives, we finetune the Stable Diffusion model on a subset of COCO image-caption pairs \cite{coco}. Specifically, we first select all unique images from the Visual Spatial Reasoning \cite{vsr_dataset} dataset, as these images have fewer visual-linguistic ambiguities and a greater number of different categories that appear in each image. Then, we use a CLIP model \cite{clip} to select the caption with the highest semantic similarity with respect to its corresponding image. Finally, we adopt a pretrained noun parser to parse all nouns from the captions, and leverage Grounded-SAM to generate binary segmentation masks for each noun (or noun phrase) \cite{flair, sam, grounding_dino}. The final dataset consists of $\sim$ 4526 image-caption pairs and their respective binary segmentation masks. We illustrate a high-level data and training pipeline in Figure \ref{fig:method}.

\textbf{Setup.} Our main experiments are performed using Stable Diffusion v1.4 \cite{ldm}, a popular text-to-image diffusion model for high-quality generation. We use both $\mathcal{L}_\text{token}$ and $\mathcal{L}_\text{pixel}$ in addition to its original denoising objective $\lsimplelcm$ with a constant global learning rate of 5e-6 for 24,000 steps (32,000 for Stable Diffusion v2.1) using the AdamW optimizer \cite{adamw}. We trained the entire U-Net with a batch size of 1 and 4 gradient accumulation steps on a single GPU. We apply center crop to all training images and their respective segmentation maps $\mathcal{M}$. For Stable Diffusion v1.4, the cross-attention layers are located in the U-Net encoder $U_\text{E}$, the middle block $U_\text{Mid}$, and the decoder $U_\text{D}$. We apply $\mathcal{L}_\text{token}$ and $\mathcal{L}_\text{pixel}$ to all cross-attention layers in $U_\text{Mid}$ and $U_\text{D}$.

\subsection{Main Results}

\textbf{Baselines.} We compare our finetuned Stable Diffusion v1.4 model against several baselines: (1) \textbf{Composable Diffusion} \cite{composable_diffusion}, which decomposes the prompt into different conditions and uses a score-based mechanism to denoise the image; (2) \textbf{Layout Guidance Diffusion} \cite{layout_diffusion}, which backpropagates cross-attention map gradients to the noisy latent with user-specified object tokens and bounding boxes for spatially controllable compositional generation; (3) \textbf{Structured Diffusion} \cite{structured_diffusion}, which automatically parses the prompt into a constituency tree and manipulates cross-attention key and value for compositional generation; and (4) \textbf{Attend-and-Excite} \cite{attend_and_excite}, which applies Gaussian kernels attention maps from user-specified tokens, and uses smoothed attention maps for compositional generation.

\textbf{Multi-category Instance Composition.} Benchmarks that examine this compositionality focus on the successful generation of multiple categories of instances in the image mentioned in the text condition. We use an existing benchmark, VISOR \cite{visor}, along with our benchmark, \textsc{MultiGen}, to study the model's capability in multi-category instance composition. The VISOR benchmark obtains all unique pairwise combinations of the 80 object categories from COCO \cite{coco} and converts each pair (A, B) into a text prompt with an arbitrary spatial relationship (R) following a template ``$<$A$>$ $<$R$>$ $<$B$>$", for example, ``a motorcycle to the left of an elephant." With the images generated from such prompts, VISOR uses an open-vocabulary detector \cite{owl} to detect the presence and spatial locations of each category in the pair. Object Accuracy (OA) measures the successful rate of generating instances from both categories. VISOR also provides metrics for spatial relationships. However, since our work focuses on multi-category instance composition, we only adopt the OA metric, and report relevant numbers in Table \ref{tab:main_result}.

\begin{table*}[ht]
\centering
\scalebox{0.66}{

\begin{tabular}{llll|ccccccccccc}
\hline
\toprule
 \multicolumn{1}{l}{} & \multicolumn{3}{c}{\textbf{}} & \multicolumn{1}{c}{} & \phantom{} & \multicolumn{3}{c}{\textsc{COCO Instances} ($\uparrow$)} & \phantom{} & \multicolumn{3}{c}{\textsc{ADE20K Instances} ($\uparrow$)} & \phantom{} & \multicolumn{1}{c}{} \\

 \cmidrule{7-9} \cmidrule{11-13} 

 \multicolumn{1}{l}{} & \multicolumn{1}{l}{\textbf{Component}} & \multicolumn{1}{l}{\textbf{Original}} & \multicolumn{1}{l|}{\textbf{Modified}} & \multicolumn{1}{c}{OA ($\uparrow$)} & \phantom{} & \multicolumn{1}{l}{MG3} & \multicolumn{1}{l}{MG4} & \multicolumn{1}{l}{MG5} & \phantom{} & \multicolumn{1}{l}{MG3} & \multicolumn{1}{l}{MG4} & \multicolumn{1}{l}{MG5} & \phantom{}  & \multicolumn{1}{c}{FID (\textbf{C}) ($\downarrow$)} \\

\midrule
 \cellcolorlightgreen & \multicolumn{3}{c|}{\cellcolorlightgreen \emph{\textbf{Ours (Stable Diffusion 1.4)}}} & \textbf{52.15} & \phantom{} & \textbf{76.16}\textsubscript{1.04} & \textbf{28.81}\textsubscript{0.95} & \textbf{3.28}\textsubscript{0.48} & \phantom{} & \textbf{76.93}\textsubscript{1.09} & \textbf{33.92}\textsubscript{1.47} & \textbf{6.21}\textsubscript{0.62} & \phantom{} & \textbf{20.19} \\
 
\midrule
\midrule
\cellcolorlightblue \hypertarget{ab1}{\textbf{(i)}} & \cellcolorlightblue finetune & \cellcolorlightblue $\lsimplelcm + \lambda \mathcal{L}_{\text{token}} + \gamma \mathcal{L}_{\text{pixel}}$ & \cellcolorlightblue frozen & 29.86 & \phantom{} & 50.74\textsubscript{0.89} & 11.68\textsubscript{0.45} & 0.88\textsubscript{0.21} & \phantom{} & 53.96\textsubscript{1.14} & 16.52\textsubscript{1.13} & 1.89\textsubscript{0.34} & \phantom{} & 20.88 \\

\midrule
\cellcolorlightgray \hypertarget{ab2}{\textbf{(ii)}} & \cellcolorlightgray $\lambda \mathcal{L}_{\text{token}} + \gamma \mathcal{L}_{\text{pixel}}$ & \cellcolorlightgray $\lsimplelcm + \lambda \mathcal{L}_{\text{token}} + \gamma \mathcal{L}_{\text{pixel}}$ & \cellcolorlightgray $\lsimplelcm$ & 38.02 & \phantom{} & 63.21\textsubscript{1.73} & 19.03\textsubscript{1.28} & 1.88\textsubscript{0.23} & \phantom{} & 62.86\textsubscript{1.41} & 22.17\textsubscript{1.14} & 3.27\textsubscript{0.46} & \phantom{} & 23.04 \\

\midrule
\cellcolorlightpurple \hypertarget{ab3}{\textbf{(iii)}} & \cellcolorlightpurple $\lambda \mathcal{L}_{\text{token}}$ & \cellcolorlightpurple $\lsimplelcm + \lambda \mathcal{L}_{\text{token}} + \gamma \mathcal{L}_{\text{pixel}}$ & \cellcolorlightpurple $\lsimplelcm + \gamma \mathcal{L}_{\text{pixel}}$ & 37.46 & \phantom{} & 61.95\textsubscript{1.05} & 18.44\textsubscript{0.98} & 1.81\textsubscript{0.35} & \phantom{} & 65.11\textsubscript{0.99} & 25.34\textsubscript{0.95} & 4.18\textsubscript{0.53} & \phantom{} & 22.32 \\

\midrule
\cellcolorlightpurple \hypertarget{ab4}{\textbf{(iv)}} & \cellcolorlightpurple $\gamma \mathcal{L}_{\text{pixel}}$ & \cellcolorlightpurple $\lsimplelcm + \lambda \mathcal{L}_{\text{token}} + \gamma \mathcal{L}_{\text{pixel}}$ & \cellcolorlightpurple $\lsimplelcm +\lambda \mathcal{L}_{\text{token}}$ & 49.85 & \phantom{} & 71.61\textsubscript{1.06} & 24.94\textsubscript{1.24} & \uline{2.83}\textsubscript{0.62} & \phantom{} & 75.37\textsubscript{0.90} & 32.91\textsubscript{1.53} & 5.58\textsubscript{0.46} & \phantom{} & 20.60 \\

\midrule
    & & & $U_{E}^{16\times16}$, $U_{\text{Mid}}^{8 \times 8}$, $U_{D}^{16\times16}$ & 42.92 & \phantom{} & 64.42\textsubscript{1.08} & 18.78\textsubscript{1.03} & 1.47\textsubscript{0.32} & \phantom{} & 67.24\textsubscript{0.98} & 24.77\textsubscript{1.01} & 3.60\textsubscript{0.57} & \phantom{} & \uline{20.45} \\[2pt]
    
    & layers w. & $U_{\text{Mid}}^{8 \times 8}$, $U_{D}^{16\times16}$, & $U_{D}^{16\times16}$ & 41.66 & \phantom{} & 66.17\textsubscript{1.29} & 20.29\textsubscript{1.20} & 1.98\textsubscript{0.48} & \phantom{} & 67.53\textsubscript{1.03} & 25.45\textsubscript{1.24} & 3.87\textsubscript{0.45} & \phantom{} & 20.66 \\[2pt]
    
    & $\lambda \mathcal{L}_{\text{token}} + \gamma  \mathcal{L}_{\text{pixel}}$ & $U_{D}^{32\times32}$, $U_{D}^{64\times64}$ & $U_{\text{Mid}}^{8 \times 8}$, $U_{D}^{16\times16}$ & 44.27 & \phantom{} & 65.47\textsubscript{1.42} & 19.16\textsubscript{1.07} & 1.59\textsubscript{0.34} & \phantom{} & 69.33\textsubscript{1.33} & 26.80\textsubscript{1.52} & 4.05\textsubscript{0.68} & \phantom{} & 20.75 \\[2pt]

    \multirow{-4}{*}{\hypertarget{ab5}{\textbf{(v)}}} & &  & 
    $U_{\text{Mid}}^{8 \times 8}$, $U_{D}^{16\times16}$, $U_{D}^{32\times32}$ & \uline{49.89} & \phantom{} & \uline{73.80}\textsubscript{1.33} & \uline{26.28}\textsubscript{1.00} & 2.76\textsubscript{0.35} & \phantom{} & \uline{75.49}\textsubscript{1.02} & \uline{33.47}\textsubscript{1.27} & \uline{5.87}\textsubscript{0.80} & \phantom{} & \uline{20.45} \\[2pt]
\bottomrule
\hline
\end{tabular}
}
\vspace{-1ex}
\caption{\textbf{Ablation studies}. We show how different objectives and cross-attention layers with $\mathcal{L}_{\text{token}}$ and $\mathcal{L}_{\text{pixel}}$ affect the multi-category instance compositon and photorealism. Qualitative visualizations of the cross-attention maps for (i) - (iv) are in Figure \ref{fig:activation_viz}.}
\label{tab:ablation}
\vspace{-4ex}
\end{table*}

\textsc{MultiGen} uses a similar evaluation strategy compared to VISOR, but is designed to be a more challenging metric for multi-category instance composition. Specifically, given a set of distinct instance categories of size $N$, we randomly sample 5 categories (\textit{e.g,} A, B, C, D, E), format them into a sentence (\textit{i.e.,} A photo of A, B, C, D, and E.), and use them as the condition for a text-to-image diffusion model to generate the image. Then, we use a strong open-vocabulary detector \cite{owlv2} to detect the presence of these categories in the generated image. We perform the sampling process for 80 categories of COCO instances \cite{coco} and 100 categories of ADE20K instances \cite{ade20k_scene, ade20k_semantic} 1,000 times, resulting in 1,000 text prompts as multi-category instance combinations from each dataset. For each generated image, we leverage the detector to detect how many categories of instances appear in the image. We aggregate the overall success rate of generating 2-5 specified categories out of 5 as MG2-5.

Compositional image generation often involves inference variance. To account for this, each prompt in \textsc{MultiGen} is used to generate 10 rounds of images, which results in 10 $\times$ 1000 images being generated for each dataset's category combinations. We calculate MG2-5 for each round and report the mean and standard deviation (in \textsubscript{subscript}) of the MG2-5 success rate out of 10 rounds in Table \ref{tab:main_result}. Based on the evaluation, our model exceeds all baselines in object accuracy and MG2-4. We find that Attend-and-Excite \cite{attend_and_excite} has a considerable success rate in generating all 5 categories, but it falls short of generating 2-4 categories comparably. We conjecture that this is due to the training distribution where captions do not include as many as 5 or more categories of instances, which inevitably leads to diminishing improvements in multi-category instance composition.

\begin{figure*}[ht!]
  \centering
    \includegraphics[width=1\linewidth]{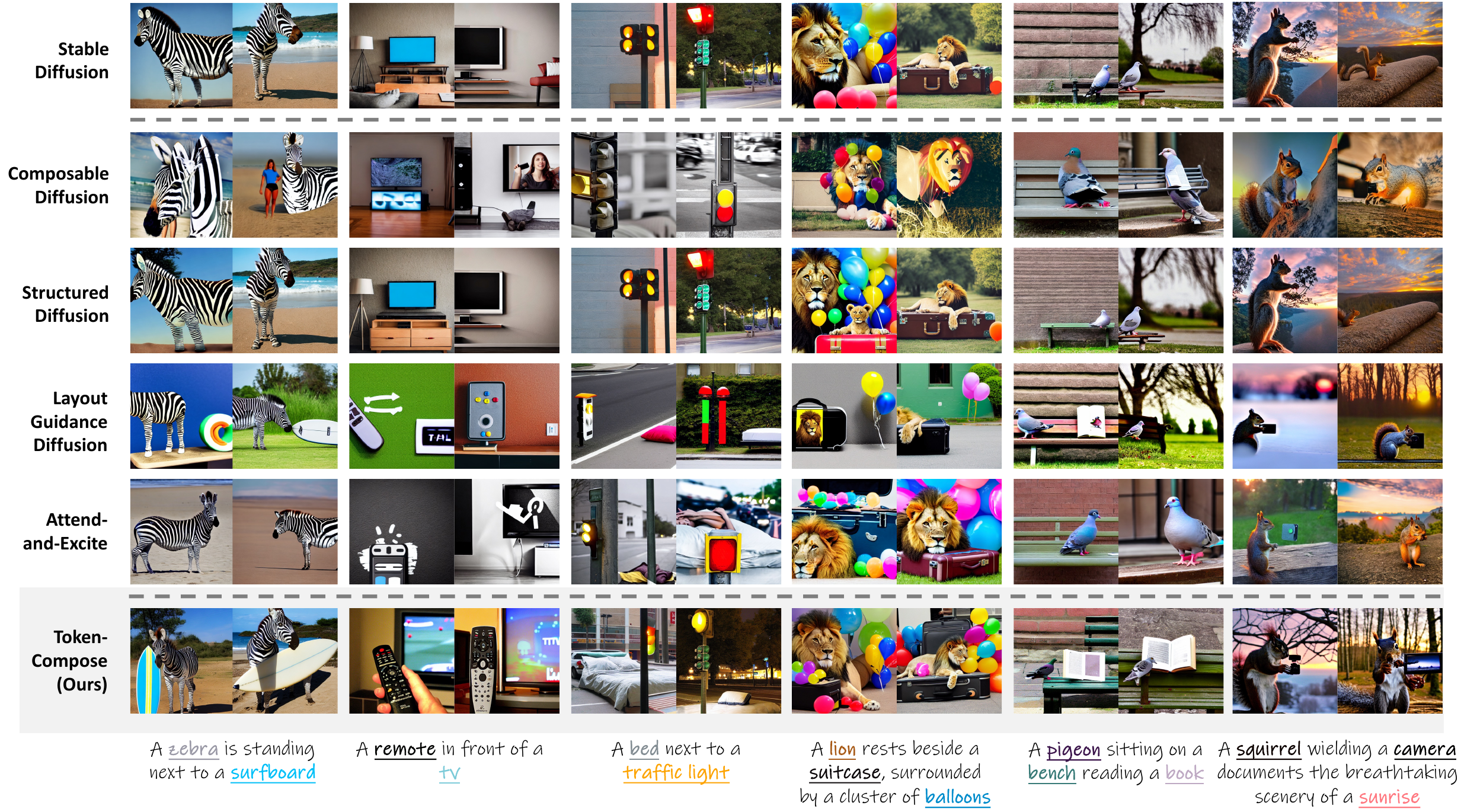}
    \vspace{-5ex}
    \caption{\textbf{Qualitative comparison between baselines and our model.} We demonstrate the effectiveness of our training framework in multi-category instance composition compared with a frozen \textbf{Stable Diffusion} Model \cite{ldm}, \textbf{Composable Diffusion} \cite{composable_diffusion}, \textbf{Structured Diffusion} \cite{structured_diffusion}, \textbf{Layout Guidance Diffusion} \cite{layout_diffusion}, and \textbf{Attend-and-Excite} \cite{attend_and_excite}. The first three columns show composition of two categories that is deemed difficult to be generated from a pretrained Stable Diffusion model (due to rare chances of co-occurrence or significant difference in instance sizes in the real world). The last three columns show the composition of three categories where composing them requires understanding of visual representations of each text token.}
    \label{fig:comparisons}
\vspace{-4ex}
\end{figure*}

\textbf{Photorealism.} We compare the image quality generated from the baselines and our model using the Fréchet Inception Distance (FID) metric \cite{fid}. For baselines that do not require additional conditional input other than captions, we calculate the FID score based on 10,000 image-caption pairs sampled from the \textbf{C}OCO validation set (\textbf{C}). We also report the FID metric based on 1,000 image-caption pairs from the validation set of \textbf{F}lickr30K entities (\textbf{F}) \cite{flickr30k}. As this dataset provides labels and bounding boxes for entities in the captions \cite{flickr30k_anno}, it can be used by inference-based methods that require such input. We report all applicable scores in Table \ref{tab:main_result}. We also qualitatively evaluate image quality in conjunction with multi-category instance composition across different baselines compared to our model in Figure \ref{fig:comparisons}. For fairness, we use the same initial latent in each comparison.

\textbf{Efficiency.} As a training-based method, our model does not require additional inference-time manipulations compared to a standard text-to-image diffusion pipeline. On the other hand, the majority of inference-based methods impose a non-ignorable compute burden for compositional generation, where the slowest baseline, Attend-and-Excite \cite{attend_and_excite}, takes more than 3$\times$ of time to generate a single image. We report efficiency results in Table \ref{tab:main_result} in seconds needed to generate an image on a single NVIDIA RTX 3090 GPU with 50 DDIM steps \cite{ddim} and classifier-free guidance \cite{classifier_free_guidance}.

\vspace{-1ex}
\subsection{Generalization}
We evaluate whether our training method generalizes to different variants of text-to-image models. To this end, we apply $\mathcal{L}_{\text{token}}$ and $\mathcal{L}_{\text{pixel}}$ in addition to $\lsimplelcm$ to Stable Diffusion v2.1, and compare the results in multi-category instance composition and photorealism between a frozen baseline and a baseline trained only with $\lsimplelcm$ in Table \ref{tab:sd21_table}. The results show that these grounding objectives also benefit Stable Diffusion v2.1 remarkably.

\begin{table}[hb]
\vspace{-3ex}
\centering
\scalebox{0.65}{
\begin{tabular}{lc|ccc|cccccc@{}}
\hline
\toprule

\multicolumn{1}{c}{\textbf{}} & \multicolumn{7}{c}{\textbf{Multi-category Instance Composition $(\uparrow)$}} &  \multicolumn{1}{c}{} \\

\cmidrule{2-8}

\textbf{Model} & & \multicolumn{3}{c|}{\textsc{COCO Instances}}  & \multicolumn{3}{c}{\textsc{ADE20K Instances}}  & FID $(\downarrow)$ \\

 & OA & MG3 & MG4 & MG5 & MG3 & MG4 & MG5   & (\textbf{C})\\

\midrule
\textbf{\emph{SD 2.1}} & & & & & & & & \\

\quad frozen & 47.82 & 70.14 & 25.57 & 3.27 & 75.13 & 35.07 & 7.16  & 19.59 \\

\quad ft. w. \small $\lsimplelcm$ & 55.09 & 76.43 & 32.07 & 4.73 & 77.60 & 37.09 & 7.78  & 20.55 \\

\midrule
\colorcello \quad \emph{\textbf{Ours}} & \colorcello \textbf{60.10} & \colorcello \textbf{80.48} & \colorcello \textbf{36.69} & \colorcello \textbf{5.71} & \colorcello \textbf{79.51} & \colorcello \textbf{39.59} & \colorcello \textbf{8.13} &  \colorcello \textbf{19.15} \\

\bottomrule
\hline
\end{tabular}
}
\vspace{-1ex}
\caption{\textbf{Model generalization.} We show multi-category instance composition and FID results when we apply our training approach to Stable Diffusion v2.1, which has a different (1) input and cross-attention map resolution, (2) text encoder \cite{openclip}, and (3) training schema (\textit{i.e.,} progressive distillation) \cite{v-obj}.} 
\label{tab:sd21_table}
\vspace{-2ex}
\end{table}

\subsection{Knowledge Transfer}
By learning segmentation maps of each noun token via cross-attention with $\mathcal{L}_{\text{token}}$ and $\mathcal{L}_{\text{pixel}}$, the model is expected to gain enhanced abilities to segment instances in the image. To verify that this knowledge is transferred from a segmentation model \cite{sam} to a diffusion model, we leverage the COCO-Gen dataset from DAAM \cite{daam}. For a fair comparison, we performed null text inversion \cite{null_text_inversion} on all images from this dataset for each model. As the model reconstructs the images, we use DAAM's algorithm and evaluation protocol to calculate the mIoU between the model's cross-attention map and human-annotated segmentation maps. The results are shown in Table \ref{tab:segmentation}.

\begin{SCtable}[1.5][hb!]
\caption{\textbf{Enhanced segmentation capabilities.} We examine the grounded segmentaion knowledge transfer from Grounded-SAM \cite{sam, grounding_dino} to our finetuned diffusion model with DAAM \cite{daam}.}
\label{tab:segmentation}
\scalebox{0.77}{

\begin{tabular}{l|c}
\hline
\toprule
\multicolumn{1}{l}{\multirow{-1}{*}{\textbf{Model}}} & \multicolumn{1}{c}{mIoU $(\uparrow)$}  \\
\midrule
\textbf{\emph{SD 1.4}} \\
\quad frozen & 0.5371 \\
\quad ft. w. \small $\lsimplelcm$ & 0.5412 \\
\midrule
\colorcello \quad \emph{\textbf{Ours}} & \colorcello \textbf{0.5876} \\
\bottomrule
\hline
\end{tabular}
}

\end{SCtable}
\vspace{-1ex}
\subsection{Downstream Metrics}
While \textsc{TokenCompose} is not explicitly optimized for binding attributes such as color, texture, shape to or specifying relations between objects, we show that, by improving the model's capability in multi-category instance composition, we also observe quantitative improvements in downstream compositionality metrics. We evaluate our model using benchmarks proposed by T2I-CompBench \cite{t2i_bench}, which uses various expert models \cite{blip, clip, unidet} to judge the alignment between compositional prompt and the generated image. We show results of this benchmark in Table \ref{tab:additional_benefits}.

We believe that improvements in these metrics are due to the model having higher chances of composing multiple categories of instances, which serves as a prerequisite for assigning attributes to and relations between objects.

\begin{table}[hb!]
\centering
\scalebox{0.69}{
\begin{tabular}{l|ccccccc}
\hline
\toprule
 \multicolumn{1}{l}{} & \multicolumn{3}{c}{\textbf{Attribute Binding} ($\uparrow$)}  & & \multicolumn{3}{c}{\textbf{Object Relations ($\uparrow$)}}\\
 \cmidrule{2-4} \cmidrule{6-8}
 
\multicolumn{1}{l}{\multirow{-3}{*}{\textbf{Model}}} & \small Color & \small Shape & \small Texture & \phantom{} & \small Spatial & \small Non-Spat. & \small Complex \\

\midrule
\textbf{\emph{SD 1.4}} \\
\quad frozen & 0.3765 & 0.3576 & 0.4642 & \phantom{} & 0.1161 & 0.3102 & 0.2795  \\
\quad ft. w. \small $\lsimplelcm$ & 0.4647 & 0.4598 & 0.5209 & \phantom{} & 0.1326 & 0.3172 & 0.2912 \\
\midrule
\colorcello \quad \emph{\textbf{Ours}} & \colorcello \textbf{0.5055} & \colorcello \textbf{0.4852} & \colorcello \textbf{0.5881} & \colorcello \phantom{} & \colorcello \textbf{0.1815} & \colorcello \textbf{0.3173} & \colorcello \textbf{0.2937} \\

\bottomrule
\hline
\end{tabular}
}
\caption{\textbf{Improvements in downstream metrics.} As successful composition of multiple categories of instances serves as the \emph{foundation} for attribute binding and object relationship specification in compositional generation, our model shows quantitative improvement on relevant metrics, despite it is \emph{not} optimized explicitly for these downstream metrics.}
\label{tab:additional_benefits}
\vspace{-4ex}
\end{table}

\section{Ablations}

We ablate (1) incorporating different grounding objectives; and (2) layers applied with grounding objectives in training the denoising U-Net, and evaluate how different design strategies affect multi-category instance composition and photorealism. We show our ablation results in Table \ref{tab:ablation}.

\subsection{Grounding Objectives}
We compare our model with a model trained only with $\lsimplelcm$ for the same number of optimization steps in (\hyperlink{ab2}{ii}). We find that there is a moderate improvement in metrics related to multi-category instance composition comparing to (\hyperlink{ab1}{i}), followed by a degeneration in photorealism (\textit{i.e.,} FID). We conjecture that training a model only with $\lsimplelcm$ in the COCO dataset brings an inherent advantage in improving composing multiple categories of instances, as the dataset often contains image-caption pairs where multiple categories of instances appear in the image and are encapsulated by the caption.

We then trained a model with only $\lsimplelcm$ and $\mathcal{L}_{\text{pixel}}$ (\hyperlink{ab3}{iii}), and a model trained  with only $\lsimplelcm$ and $\mathcal{L}_{\text{token}}$ (\hyperlink{ab4}{iv}). We observe that $\mathcal{L}_{\text{pixel}}$ has little effect compared to $\mathcal{L}_{\text{token}}$ when used \emph{alone} with $\lsimplelcm$. $\mathcal{L}_{\text{token}}$ plays a major role in improving multi-category instance composition and photorealism. However, empirically, we find that training the model with only $\lsimplelcm$ and $\mathcal{L}_{\text{token}}$ often leads to attention maps that overly activate in subregions of the instance (see Figure \ref{fig:activation_viz}). We also observe that different training runs on this combination of losses lead to unstable inference performance in multi-category instance composition.
\vspace{-1ex}
\subsection{Layers with Grounding Objectives}
Finally, we experiment with adding $\mathcal{L}_{\text{pixel}}$ and $\mathcal{L}_{\text{token}}$ at different layers of cross-attention of the denoising U-Net (\hyperlink{ab5}{v}). We find that adding grounding objectives to the middle block and the decoder of the U-Net improves the overall performance of multi-category instance composition. Removing the constraint from the middle block or adding the constraint to the encoder degrades the performance. Furthermore, for cross-attention layers at the decoder with variable resolutions, we find that the more layers optimized with $\mathcal{L}_{\text{pixel}}$ and $\mathcal{L}_{\text{token}}$, the better the performance in both multi-category instance composition and photorealism.

\vspace{-1ex}
\section{Limitation \& Conclusion}
\textbf{Limitation.} As one of the pioneering works exploring the potential to improve a text-conditioned generative model with image-token consistency using an understanding model, we only add supervision terms to noun tokens for the text prompt. While we show that this approach improves multi-category instance composition significantly, there are many more elements from the text prompts that one can leverage an understanding model to improve a generative model, such as adjectives, verbs, and/or determiners as fine-grained token-level training objectives.

\textbf{Conclusion.} We explore the possibility of leveraging foundation image understanding models to improve grounding capabilities of a text-conditioned generative model. Our training framework, \textsc{TokenCompose}, excels at multi-category instance composition with improved image quality. To facilitate research in this niche, we also propose \textsc{MultiGen}, a challenging benchmark that requires a model to generate multiple categories of instances in one image. As a fundamental challenge in compositional generation, we hope our training framework and benchmark can inspire future works that effectively leverage the synergy between understanding and generation to improve either or both directions.
\vspace{-2ex}
\section*{Acknowledgement}
This work is supported by NSF Award IIS-2127544. We thank Yifan Xu and Dou Kwark from UC San Diego, Kaiyi Huang from University of Hong Kong, and Adithya Bhaskar and Ofir Press from the Princeton NLP Group for discussions and/or feedback.

{
    \small
    \bibliographystyle{ieeenat_fullname}
    \bibliography{main}

\begin{thebibliography}{84}
\providecommand{\natexlab}[1]{#1}
\providecommand{\url}[1]{\texttt{#1}}
\expandafter\ifx\csname urlstyle\endcsname\relax
  \providecommand{\doi}[1]{doi: #1}\else
  \providecommand{\doi}{doi: \begingroup \urlstyle{rm}\Url}\fi

\bibitem[Akbik et~al.(2019)Akbik, Bergmann, Blythe, Rasul, Schweter, and Vollgraf]{flair}
Alan Akbik, Tanja Bergmann, Duncan Blythe, Kashif Rasul, Stefan Schweter, and Roland Vollgraf.
\newblock {FLAIR}: An easy-to-use framework for state-of-the-art {NLP}.
\newblock In \emph{{NAACL} 2019, 2019 Annual Conference of the North American Chapter of the Association for Computational Linguistics (Demonstrations)}, pages 54--59, 2019.

\bibitem[Bakr et~al.(2023)Bakr, Sun, Shen, Khan, Li, and Elhoseiny]{hrs_bench}
Eslam~Mohamed Bakr, Pengzhan Sun, Xiaogian Shen, Faizan~Farooq Khan, Li~Erran Li, and Mohamed Elhoseiny.
\newblock Hrs-bench: Holistic, reliable and scalable benchmark for text-to-image models.
\newblock In \emph{Proceedings of the IEEE/CVF International Conference on Computer Vision}, pages 20041--20053, 2023.

\bibitem[Betker et~al.(2023)Betker, Goh, Jing, TimBrooks, Wang, Li, LongOuyang, JuntangZhuang, JoyceLee, YufeiGuo, WesamManassra, PrafullaDhariwal, CaseyChu, YunxinJiao, and Ramesh]{dalle3}
James Betker, Gabriel Goh, Li Jing, † TimBrooks, Jianfeng Wang, Linjie Li, † LongOuyang, † JuntangZhuang, † JoyceLee, † YufeiGuo, † WesamManassra, † PrafullaDhariwal, † CaseyChu, † YunxinJiao, and Aditya Ramesh.
\newblock Improving image generation with better captions, 2023.

\bibitem[Black et~al.(2023)Black, Janner, Du, Kostrikov, and Levine]{diffusion_rl}
Kevin Black, Michael Janner, Yilun Du, Ilya Kostrikov, and Sergey Levine.
\newblock Training diffusion models with reinforcement learning.
\newblock In \emph{ICML 2023 Workshop on Structured Probabilistic Inference {\&} Generative Modeling}, 2023.

\bibitem[Caron et~al.(2021)Caron, Touvron, Misra, J{\'e}gou, Mairal, Bojanowski, and Joulin]{dino}
Mathilde Caron, Hugo Touvron, Ishan Misra, Herv{\'e} J{\'e}gou, Julien Mairal, Piotr Bojanowski, and Armand Joulin.
\newblock Emerging properties in self-supervised vision transformers.
\newblock In \emph{Proceedings of the IEEE/CVF international conference on computer vision}, pages 9650--9660, 2021.

\bibitem[Chefer et~al.(2023)Chefer, Alaluf, Vinker, Wolf, and Cohen-Or]{attend_and_excite}
Hila Chefer, Yuval Alaluf, Yael Vinker, Lior Wolf, and Daniel Cohen-Or.
\newblock Attend-and-excite: Attention-based semantic guidance for text-to-image diffusion models.
\newblock \emph{ACM Transactions on Graphics (TOG)}, 42\penalty0 (4):\penalty0 1--10, 2023.

\bibitem[Chen et~al.(2024)Chen, YU, GE, Yao, Xie, Wang, Kwok, Luo, Lu, and Li]{chen2024pixartalpha}
Junsong Chen, Jincheng YU, Chongjian GE, Lewei Yao, Enze Xie, Zhongdao Wang, James Kwok, Ping Luo, Huchuan Lu, and Zhenguo Li.
\newblock Pixart-\${\textbackslash}alpha\$: Fast training of diffusion transformer for photorealistic text-to-image synthesis.
\newblock In \emph{The Twelfth International Conference on Learning Representations}, 2024.

\bibitem[Chen et~al.(2023{\natexlab{a}})Chen, Laina, and Vedaldi]{layout_diffusion}
Minghao Chen, Iro Laina, and Andrea Vedaldi.
\newblock Training-free layout control with cross-attention guidance.
\newblock \emph{arXiv preprint arXiv:2304.03373}, 2023{\natexlab{a}}.

\bibitem[Chen et~al.(2023{\natexlab{b}})Chen, Sun, Song, and Luo]{diffusiondet}
Shoufa Chen, Peize Sun, Yibing Song, and Ping Luo.
\newblock Diffusiondet: Diffusion model for object detection.
\newblock In \emph{Proceedings of the IEEE/CVF International Conference on Computer Vision}, pages 19830--19843, 2023{\natexlab{b}}.

\bibitem[Cheng et~al.(2022)Cheng, Misra, Schwing, Kirillov, and Girdhar]{mask2former}
Bowen Cheng, Ishan Misra, Alexander~G Schwing, Alexander Kirillov, and Rohit Girdhar.
\newblock Masked-attention mask transformer for universal image segmentation.
\newblock In \emph{Proceedings of the IEEE/CVF conference on computer vision and pattern recognition}, pages 1290--1299, 2022.

\bibitem[Cho et~al.(2023)Cho, Zala, and Bansal]{paintskills}
Jaemin Cho, Abhay Zala, and Mohit Bansal.
\newblock Dall-eval: Probing the reasoning skills and social biases of text-to-image generation models.
\newblock In \emph{Proceedings of the IEEE/CVF International Conference on Computer Vision}, pages 3043--3054, 2023.

\bibitem[Clark and Jaini(2023)]{diffusion_zs_classification}
Kevin Clark and Priyank Jaini.
\newblock Text-to-image diffusion models are zero-shot classifiers.
\newblock In \emph{ICLR 2023 Workshop on Mathematical and Empirical Understanding of Foundation Models}, 2023.

\bibitem[Dayma et~al.(2021)Dayma, Patil, Cuenca, Saifullah, Abraham, Lê~Khắc, Melas, and Ghosh]{dalle_mini}
Boris Dayma, Suraj Patil, Pedro Cuenca, Khalid Saifullah, Tanishq Abraham, Phúc Lê~Khắc, Luke Melas, and Ritobrata Ghosh.
\newblock Dall·e mini, 2021.

\bibitem[Ding et~al.(2021)Ding, Yang, Hong, Zheng, Zhou, Yin, Lin, Zou, Shao, Yang, et~al.]{cogview}
Ming Ding, Zhuoyi Yang, Wenyi Hong, Wendi Zheng, Chang Zhou, Da Yin, Junyang Lin, Xu Zou, Zhou Shao, Hongxia Yang, et~al.
\newblock Cogview: Mastering text-to-image generation via transformers.
\newblock \emph{Advances in Neural Information Processing Systems}, 34:\penalty0 19822--19835, 2021.

\bibitem[Ding et~al.(2022)Ding, Zheng, Hong, and Tang]{cogview2}
Ming Ding, Wendi Zheng, Wenyi Hong, and Jie Tang.
\newblock Cogview2: Faster and better text-to-image generation via hierarchical transformers.
\newblock \emph{Advances in Neural Information Processing Systems}, 35:\penalty0 16890--16902, 2022.

\bibitem[Dosovitskiy et~al.(2021)Dosovitskiy, Beyer, Kolesnikov, Weissenborn, Zhai, Unterthiner, Dehghani, Minderer, Heigold, Gelly, Uszkoreit, and Houlsby]{vit}
Alexey Dosovitskiy, Lucas Beyer, Alexander Kolesnikov, Dirk Weissenborn, Xiaohua Zhai, Thomas Unterthiner, Mostafa Dehghani, Matthias Minderer, Georg Heigold, Sylvain Gelly, Jakob Uszkoreit, and Neil Houlsby.
\newblock An image is worth 16x16 words: Transformers for image recognition at scale.
\newblock In \emph{International Conference on Learning Representations}, 2021.

\bibitem[Feng et~al.(2023{\natexlab{a}})Feng, He, Fu, Jampani, Akula, Narayana, Basu, Wang, and Wang]{structured_diffusion}
Weixi Feng, Xuehai He, Tsu-Jui Fu, Varun Jampani, Arjun~Reddy Akula, Pradyumna Narayana, Sugato Basu, Xin~Eric Wang, and William~Yang Wang.
\newblock Training-free structured diffusion guidance for compositional text-to-image synthesis.
\newblock In \emph{The Eleventh International Conference on Learning Representations}, 2023{\natexlab{a}}.

\bibitem[Feng et~al.(2023{\natexlab{b}})Feng, Zhang, Yu, Fang, Li, Chen, Lu, Liu, Yin, Feng, et~al.]{feng2023ernie}
Zhida Feng, Zhenyu Zhang, Xintong Yu, Yewei Fang, Lanxin Li, Xuyi Chen, Yuxiang Lu, Jiaxiang Liu, Weichong Yin, Shikun Feng, et~al.
\newblock Ernie-vilg 2.0: Improving text-to-image diffusion model with knowledge-enhanced mixture-of-denoising-experts.
\newblock In \emph{Proceedings of the IEEE/CVF Conference on Computer Vision and Pattern Recognition}, pages 10135--10145, 2023{\natexlab{b}}.

\bibitem[Gokhale et~al.(2022)Gokhale, Palangi, Nushi, Vineet, Horvitz, Kamar, Baral, and Yang]{visor}
Tejas Gokhale, Hamid Palangi, Besmira Nushi, Vibhav Vineet, Eric Horvitz, Ece Kamar, Chitta Baral, and Yezhou Yang.
\newblock Benchmarking spatial relationships in text-to-image generation.
\newblock \emph{arXiv preprint arXiv:2212.10015}, 2022.

\bibitem[Goodfellow et~al.(2014)Goodfellow, Pouget-Abadie, Mirza, Xu, Warde-Farley, Ozair, Courville, and Bengio]{gan}
Ian Goodfellow, Jean Pouget-Abadie, Mehdi Mirza, Bing Xu, David Warde-Farley, Sherjil Ozair, Aaron Courville, and Yoshua Bengio.
\newblock Generative adversarial nets.
\newblock In \emph{Advances in Neural Information Processing Systems}. Curran Associates, Inc., 2014.

\bibitem[Graves et~al.(2023)Graves, Srivastava, Atkinson, and Gomez]{bayesian}
Alex Graves, Rupesh~Kumar Srivastava, Timothy Atkinson, and Faustino Gomez.
\newblock Bayesian flow networks.
\newblock \emph{arXiv preprint arXiv:2308.07037}, 2023.

\bibitem[Hertz et~al.(2022)Hertz, Mokady, Tenenbaum, Aberman, Pritch, and Cohen-Or]{hertz2022prompt}
Amir Hertz, Ron Mokady, Jay Tenenbaum, Kfir Aberman, Yael Pritch, and Daniel Cohen-Or.
\newblock Prompt-to-prompt image editing with cross attention control.
\newblock \emph{arXiv preprint arXiv:2208.01626}, 2022.

\bibitem[Hessel et~al.(2021)Hessel, Holtzman, Forbes, Le~Bras, and Choi]{clip_score}
Jack Hessel, Ari Holtzman, Maxwell Forbes, Ronan Le~Bras, and Yejin Choi.
\newblock {CLIPS}core: A reference-free evaluation metric for image captioning.
\newblock In \emph{Proceedings of the 2021 Conference on Empirical Methods in Natural Language Processing}, pages 7514--7528, Online and Punta Cana, Dominican Republic, 2021. Association for Computational Linguistics.

\bibitem[Heusel et~al.(2017)Heusel, Ramsauer, Unterthiner, Nessler, and Hochreiter]{fid}
Martin Heusel, Hubert Ramsauer, Thomas Unterthiner, Bernhard Nessler, and Sepp Hochreiter.
\newblock Gans trained by a two time-scale update rule converge to a local nash equilibrium.
\newblock \emph{Advances in neural information processing systems}, 30, 2017.

\bibitem[Hinz et~al.(2020)Hinz, Heinrich, and Wermter]{soa}
Tobias Hinz, Stefan Heinrich, and Stefan Wermter.
\newblock Semantic object accuracy for generative text-to-image synthesis.
\newblock \emph{IEEE transactions on pattern analysis and machine intelligence}, 44\penalty0 (3):\penalty0 1552--1565, 2020.

\bibitem[Ho and Salimans(2021)]{classifier_free_guidance}
Jonathan Ho and Tim Salimans.
\newblock Classifier-free diffusion guidance.
\newblock In \emph{NeurIPS 2021 Workshop on Deep Generative Models and Downstream Applications}, 2021.

\bibitem[Ho et~al.(2020)Ho, Jain, and Abbeel]{ddpm}
Jonathan Ho, Ajay Jain, and Pieter Abbeel.
\newblock Denoising diffusion probabilistic models.
\newblock \emph{Advances in neural information processing systems}, 33:\penalty0 6840--6851, 2020.

\bibitem[Huang et~al.(2023)Huang, Sun, Xie, Li, and Liu]{t2i_bench}
Kaiyi Huang, Kaiyue Sun, Enze Xie, Zhenguo Li, and Xihui Liu.
\newblock T2i-compbench: A comprehensive benchmark for open-world compositional text-to-image generation.
\newblock In \emph{Thirty-seventh Conference on Neural Information Processing Systems Datasets and Benchmarks Track}, 2023.

\bibitem[Ilharco et~al.(2021)Ilharco, Wortsman, Wightman, Gordon, Carlini, Taori, Dave, Shankar, Namkoong, Miller, Hajishirzi, Farhadi, and Schmidt]{openclip}
Gabriel Ilharco, Mitchell Wortsman, Ross Wightman, Cade Gordon, Nicholas Carlini, Rohan Taori, Achal Dave, Vaishaal Shankar, Hongseok Namkoong, John Miller, Hannaneh Hajishirzi, Ali Farhadi, and Ludwig Schmidt.
\newblock Openclip, 2021.
\newblock If you use this software, please cite it as below.

\bibitem[Karras et~al.(2019)Karras, Laine, and Aila]{karras2019style}
Tero Karras, Samuli Laine, and Timo Aila.
\newblock A style-based generator architecture for generative adversarial networks.
\newblock In \emph{Proceedings of the IEEE/CVF conference on computer vision and pattern recognition}, pages 4401--4410, 2019.

\bibitem[Ke et~al.(2023)Ke, Ye, Danelljan, Liu, Tai, Tang, and Yu]{sam_hq}
Lei Ke, Mingqiao Ye, Martin Danelljan, Yifan Liu, Yu-Wing Tai, Chi-Keung Tang, and Fisher Yu.
\newblock Segment anything in high quality.
\newblock In \emph{NeurIPS}, 2023.

\bibitem[Khani et~al.(2023)Khani, Taghanaki, Sanghi, Amiri, and Hamarneh]{slime}
Aliasghar Khani, Saeid~Asgari Taghanaki, Aditya Sanghi, Ali~Mahdavi Amiri, and Ghassan Hamarneh.
\newblock Slime: Segment like me.
\newblock \emph{arXiv preprint arXiv:2309.03179}, 2023.

\bibitem[Kingma and Welling(2013)]{vae}
Diederik~P Kingma and Max Welling.
\newblock Auto-encoding variational bayes.
\newblock \emph{arXiv preprint arXiv:1312.6114}, 2013.

\bibitem[Kirillov et~al.(2023)Kirillov, Mintun, Ravi, Mao, Rolland, Gustafson, Xiao, Whitehead, Berg, Lo, Doll{\'a}r, and Girshick]{sam}
Alexander Kirillov, Eric Mintun, Nikhila Ravi, Hanzi Mao, Chloe Rolland, Laura Gustafson, Tete Xiao, Spencer Whitehead, Alexander~C. Berg, Wan-Yen Lo, Piotr Doll{\'a}r, and Ross Girshick.
\newblock Segment anything.
\newblock \emph{arXiv:2304.02643}, 2023.

\bibitem[Lee et~al.(2018)Lee, Xu, Fan, and Tu]{introspective}
Kwonjoon Lee, Weijian Xu, Fan Fan, and Zhuowen Tu.
\newblock Wasserstein introspective neural networks.
\newblock In \emph{Proceedings of the IEEE conference on computer vision and pattern recognition}, pages 3702--3711, 2018.

\bibitem[Lee et~al.(2023)Lee, Liu, Ryu, Watkins, Du, Boutilier, Abbeel, Ghavamzadeh, and Gu]{t2i_hf}
Kimin Lee, Hao Liu, Moonkyung Ryu, Olivia Watkins, Yuqing Du, Craig Boutilier, Pieter Abbeel, Mohammad Ghavamzadeh, and Shixiang~Shane Gu.
\newblock Aligning text-to-image models using human feedback.
\newblock \emph{arXiv preprint arXiv:2302.12192}, 2023.

\bibitem[Li et~al.(2023{\natexlab{a}})Li, Prabhudesai, Duggal, Brown, and Pathak]{secret_classifier}
Alexander~Cong Li, Mihir Prabhudesai, Shivam Duggal, Ellis~Langham Brown, and Deepak Pathak.
\newblock Your diffusion model is secretly a zero-shot classifier.
\newblock In \emph{ICML 2023 Workshop on Structured Probabilistic Inference {\&} Generative Modeling}, 2023{\natexlab{a}}.

\bibitem[Li et~al.(2022)Li, Li, Xiong, and Hoi]{blip}
Junnan Li, Dongxu Li, Caiming Xiong, and Steven Hoi.
\newblock Blip: Bootstrapping language-image pre-training for unified vision-language understanding and generation.
\newblock In \emph{International Conference on Machine Learning}, pages 12888--12900. PMLR, 2022.

\bibitem[Li et~al.(2023{\natexlab{b}})Li, Fischer, Ke, Ding, Danelljan, and Yu]{ovtrack}
Siyuan Li, Tobias Fischer, Lei Ke, Henghui Ding, Martin Danelljan, and Fisher Yu.
\newblock Ovtrack: Open-vocabulary multiple object tracking.
\newblock In \emph{Proceedings of the IEEE/CVF Conference on Computer Vision and Pattern Recognition}, pages 5567--5577, 2023{\natexlab{b}}.

\bibitem[Li et~al.(2023{\natexlab{c}})Li, Liu, Wu, Mu, Yang, Gao, Li, and Lee]{gligen}
Yuheng Li, Haotian Liu, Qingyang Wu, Fangzhou Mu, Jianwei Yang, Jianfeng Gao, Chunyuan Li, and Yong~Jae Lee.
\newblock Gligen: Open-set grounded text-to-image generation.
\newblock In \emph{Proceedings of the IEEE/CVF Conference on Computer Vision and Pattern Recognition}, pages 22511--22521, 2023{\natexlab{c}}.

\bibitem[Li et~al.(2023{\natexlab{d}})Li, Zhou, Zhang, Zhang, Wang, and Xie]{ov_obj_seg_diffusion}
Ziyi Li, Qinye Zhou, Xiaoyun Zhang, Ya Zhang, Yanfeng Wang, and Weidi Xie.
\newblock Open-vocabulary object segmentation with diffusion models.
\newblock In \emph{Proceedings of the IEEE/CVF International Conference on Computer Vision}, 2023{\natexlab{d}}.

\bibitem[Lin et~al.(2014)Lin, Maire, Belongie, Hays, Perona, Ramanan, Doll{\'a}r, and Zitnick]{coco}
Tsung-Yi Lin, Michael Maire, Serge Belongie, James Hays, Pietro Perona, Deva Ramanan, Piotr Doll{\'a}r, and C.~Lawrence Zitnick.
\newblock Microsoft coco: Common objects in context.
\newblock In \emph{Computer Vision -- ECCV 2014}, pages 740--755, Cham, 2014. Springer International Publishing.

\bibitem[Liu et~al.(2023{\natexlab{a}})Liu, Emerson, and Collier]{vsr_dataset}
Fangyu Liu, Guy Emerson, and Nigel Collier.
\newblock Visual spatial reasoning.
\newblock \emph{Transactions of the Association for Computational Linguistics}, 11:\penalty0 635--651, 2023{\natexlab{a}}.

\bibitem[Liu et~al.(2022)Liu, Li, Du, Torralba, and Tenenbaum]{composable_diffusion}
Nan Liu, Shuang Li, Yilun Du, Antonio Torralba, and Joshua~B Tenenbaum.
\newblock Compositional visual generation with composable diffusion models.
\newblock In \emph{European Conference on Computer Vision}, pages 423--439. Springer, 2022.

\bibitem[Liu et~al.(2023{\natexlab{b}})Liu, Zeng, Ren, Li, Zhang, Yang, Li, Yang, Su, Zhu, et~al.]{grounding_dino}
Shilong Liu, Zhaoyang Zeng, Tianhe Ren, Feng Li, Hao Zhang, Jie Yang, Chunyuan Li, Jianwei Yang, Hang Su, Jun Zhu, et~al.
\newblock Grounding dino: Marrying dino with grounded pre-training for open-set object detection.
\newblock \emph{arXiv preprint arXiv:2303.05499}, 2023{\natexlab{b}}.

\bibitem[Liu et~al.(2021)Liu, Lin, Cao, Hu, Wei, Zhang, Lin, and Guo]{swin}
Ze Liu, Yutong Lin, Yue Cao, Han Hu, Yixuan Wei, Zheng Zhang, Stephen Lin, and Baining Guo.
\newblock Swin transformer: Hierarchical vision transformer using shifted windows.
\newblock In \emph{Proceedings of the IEEE/CVF International Conference on Computer Vision (ICCV)}, 2021.

\bibitem[Loshchilov and Hutter(2019)]{adamw}
Ilya Loshchilov and Frank Hutter.
\newblock Decoupled weight decay regularization.
\newblock In \emph{International Conference on Learning Representations}, 2019.

\bibitem[Ma et~al.(2023)Ma, Yang, Ju, Zhang, Liu, Wang, Zhang, and Wang]{diffusionseg}
Chaofan Ma, Yuhuan Yang, Chen Ju, Fei Zhang, Jinxiang Liu, Yu Wang, Ya Zhang, and Yanfeng Wang.
\newblock Diffusionseg: Adapting diffusion towards unsupervised object discovery.
\newblock \emph{arXiv preprint arXiv:2303.09813}, 2023.

\bibitem[Minderer et~al.(2022)Minderer, Gritsenko, Stone, Neumann, Weissenborn, Dosovitskiy, Mahendran, Arnab, Dehghani, Shen, et~al.]{owl}
Matthias Minderer, Alexey Gritsenko, Austin Stone, Maxim Neumann, Dirk Weissenborn, Alexey Dosovitskiy, Aravindh Mahendran, Anurag Arnab, Mostafa Dehghani, Zhuoran Shen, et~al.
\newblock Simple open-vocabulary object detection.
\newblock In \emph{European Conference on Computer Vision}, pages 728--755. Springer, 2022.

\bibitem[Minderer et~al.(2023)Minderer, Gritsenko, and Houlsby]{owlv2}
Matthias Minderer, Alexey~A. Gritsenko, and Neil Houlsby.
\newblock Scaling open-vocabulary object detection.
\newblock In \emph{Thirty-seventh Conference on Neural Information Processing Systems}, 2023.

\bibitem[Mokady et~al.(2023)Mokady, Hertz, Aberman, Pritch, and Cohen-Or]{null_text_inversion}
Ron Mokady, Amir Hertz, Kfir Aberman, Yael Pritch, and Daniel Cohen-Or.
\newblock Null-text inversion for editing real images using guided diffusion models.
\newblock In \emph{Proceedings of the IEEE/CVF Conference on Computer Vision and Pattern Recognition}, pages 6038--6047, 2023.

\bibitem[Ni et~al.(2023)Ni, Zhang, Feng, Li, Guo, and Zuo]{ref_diff}
Minheng Ni, Yabo Zhang, Kailai Feng, Xiaoming Li, Yiwen Guo, and Wangmeng Zuo.
\newblock Ref-diff: Zero-shot referring image segmentation with generative models.
\newblock \emph{arXiv preprint arXiv:2308.16777}, 2023.

\bibitem[Nichol et~al.(2022)Nichol, Dhariwal, Ramesh, Shyam, Mishkin, Mcgrew, Sutskever, and Chen]{glide}
Alexander~Quinn Nichol, Prafulla Dhariwal, Aditya Ramesh, Pranav Shyam, Pamela Mishkin, Bob Mcgrew, Ilya Sutskever, and Mark Chen.
\newblock {GLIDE}: Towards photorealistic image generation and editing with text-guided diffusion models.
\newblock In \emph{Proceedings of the 39th International Conference on Machine Learning}, pages 16784--16804. PMLR, 2022.

\bibitem[Plummer et~al.(2015)Plummer, Wang, Cervantes, Caicedo, Hockenmaier, and Lazebnik]{flickr30k}
Bryan~A Plummer, Liwei Wang, Chris~M Cervantes, Juan~C Caicedo, Julia Hockenmaier, and Svetlana Lazebnik.
\newblock Flickr30k entities: Collecting region-to-phrase correspondences for richer image-to-sentence models.
\newblock In \emph{Proceedings of the IEEE international conference on computer vision}, pages 2641--2649, 2015.

\bibitem[Radford et~al.(2021)Radford, Kim, Hallacy, Ramesh, Goh, Agarwal, Sastry, Askell, Mishkin, Clark, Krueger, and Sutskever]{clip}
Alec Radford, Jong~Wook Kim, Chris Hallacy, Aditya Ramesh, Gabriel Goh, Sandhini Agarwal, Girish Sastry, Amanda Askell, Pamela Mishkin, Jack Clark, Gretchen Krueger, and Ilya Sutskever.
\newblock Learning transferable visual models from natural language supervision.
\newblock In \emph{Proceedings of the 38th International Conference on Machine Learning}, pages 8748--8763. PMLR, 2021.

\bibitem[Ramesh et~al.(2021)Ramesh, Pavlov, Goh, Gray, Voss, Radford, Chen, and Sutskever]{dalle}
Aditya Ramesh, Mikhail Pavlov, Gabriel Goh, Scott Gray, Chelsea Voss, Alec Radford, Mark Chen, and Ilya Sutskever.
\newblock Zero-shot text-to-image generation.
\newblock In \emph{International Conference on Machine Learning}, pages 8821--8831. PMLR, 2021.

\bibitem[Rassin et~al.(2023)Rassin, Hirsch, Glickman, Ravfogel, Goldberg, and Chechik]{linguistic_binding}
Royi Rassin, Eran Hirsch, Daniel Glickman, Shauli Ravfogel, Yoav Goldberg, and Gal Chechik.
\newblock Linguistic binding in diffusion models: Enhancing attribute correspondence through attention map alignment.
\newblock In \emph{Thirty-seventh Conference on Neural Information Processing Systems}, 2023.

\bibitem[Rezende and Mohamed(2015)]{norm_flow}
Danilo Rezende and Shakir Mohamed.
\newblock Variational inference with normalizing flows.
\newblock In \emph{Proceedings of the 32nd International Conference on Machine Learning}, pages 1530--1538, Lille, France, 2015. PMLR.

\bibitem[Rombach et~al.(2022)Rombach, Blattmann, Lorenz, Esser, and Ommer]{ldm}
Robin Rombach, Andreas Blattmann, Dominik Lorenz, Patrick Esser, and Bj{\"o}rn Ommer.
\newblock High-resolution image synthesis with latent diffusion models.
\newblock In \emph{Proceedings of the IEEE/CVF conference on computer vision and pattern recognition}, pages 10684--10695, 2022.

\bibitem[Ronneberger et~al.(2015)Ronneberger, Fischer, and Brox]{unet}
Olaf Ronneberger, Philipp Fischer, and Thomas Brox.
\newblock U-net: Convolutional networks for biomedical image segmentation.
\newblock In \emph{Medical Image Computing and Computer-Assisted Intervention--MICCAI 2015: 18th International Conference, Munich, Germany, October 5-9, 2015, Proceedings, Part III 18}, pages 234--241. Springer, 2015.

\bibitem[Ruiz et~al.(2023)Ruiz, Li, Jampani, Pritch, Rubinstein, and Aberman]{ruiz2023dreambooth}
Nataniel Ruiz, Yuanzhen Li, Varun Jampani, Yael Pritch, Michael Rubinstein, and Kfir Aberman.
\newblock Dreambooth: Fine tuning text-to-image diffusion models for subject-driven generation.
\newblock In \emph{Proceedings of the IEEE/CVF Conference on Computer Vision and Pattern Recognition}, pages 22500--22510, 2023.

\bibitem[Saharia et~al.(2022)Saharia, Chan, Saxena, Li, Whang, Denton, Ghasemipour, Gontijo~Lopes, Karagol~Ayan, Salimans, et~al.]{imagen}
Chitwan Saharia, William Chan, Saurabh Saxena, Lala Li, Jay Whang, Emily~L Denton, Kamyar Ghasemipour, Raphael Gontijo~Lopes, Burcu Karagol~Ayan, Tim Salimans, et~al.
\newblock Photorealistic text-to-image diffusion models with deep language understanding.
\newblock \emph{Advances in Neural Information Processing Systems}, 35:\penalty0 36479--36494, 2022.

\bibitem[Salimans and Ho(2022)]{v-obj}
Tim Salimans and Jonathan Ho.
\newblock Progressive distillation for fast sampling of diffusion models.
\newblock In \emph{International Conference on Learning Representations}, 2022.

\bibitem[Schuhmann et~al.(2022)Schuhmann, Beaumont, Vencu, Gordon, Wightman, Cherti, Coombes, Katta, Mullis, Wortsman, Schramowski, Kundurthy, Crowson, Schmidt, Kaczmarczyk, and Jitsev]{laion}
Christoph Schuhmann, Romain Beaumont, Richard Vencu, Cade~W Gordon, Ross Wightman, Mehdi Cherti, Theo Coombes, Aarush Katta, Clayton Mullis, Mitchell Wortsman, Patrick Schramowski, Srivatsa~R Kundurthy, Katherine Crowson, Ludwig Schmidt, Robert Kaczmarczyk, and Jenia Jitsev.
\newblock {LAION}-5b: An open large-scale dataset for training next generation image-text models.
\newblock In \emph{Thirty-sixth Conference on Neural Information Processing Systems Datasets and Benchmarks Track}, 2022.

\bibitem[Segalis et~al.(2023)Segalis, Valevski, Lumen, Matias, and Leviathan]{recaption}
Eyal Segalis, Dani Valevski, Danny Lumen, Yossi Matias, and Yaniv Leviathan.
\newblock A picture is worth a thousand words: Principled recaptioning improves image generation.
\newblock \emph{arXiv preprint arXiv:2310.16656}, 2023.

\bibitem[Song et~al.(2021)Song, Meng, and Ermon]{ddim}
Jiaming Song, Chenlin Meng, and Stefano Ermon.
\newblock Denoising diffusion implicit models.
\newblock In \emph{International Conference on Learning Representations}, 2021.

\bibitem[Tang et~al.(2023)Tang, Liu, Pandey, Jiang, Yang, Kumar, Stenetorp, Lin, and Ture]{daam}
Raphael Tang, Linqing Liu, Akshat Pandey, Zhiying Jiang, Gefei Yang, Karun Kumar, Pontus Stenetorp, Jimmy Lin, and Ferhan Ture.
\newblock What the {DAAM}: Interpreting stable diffusion using cross attention.
\newblock In \emph{Proceedings of the 61st Annual Meeting of the Association for Computational Linguistics (Volume 1: Long Papers)}, pages 5644--5659, Toronto, Canada, 2023. Association for Computational Linguistics.

\bibitem[Tian et~al.(2023)Tian, Aggarwal, Colaco, Kira, and Gonzalez-Franco]{diffuse_attend_segment}
Junjiao Tian, Lavisha Aggarwal, Andrea Colaco, Zsolt Kira, and Mar Gonzalez-Franco.
\newblock Diffuse, attend, and segment: Unsupervised zero-shot segmentation using stable diffusion.
\newblock \emph{arXiv preprint arXiv:2308.12469}, 2023.

\bibitem[Tong et~al.(2023)Tong, Jones, and Steinhardt]{mm_failures}
Shengbang Tong, Erik Jones, and Jacob Steinhardt.
\newblock Mass-producing failures of multimodal systems with language models.
\newblock In \emph{Advances in Neural Information Processing Systems (NeurIPS)}, 2023.

\bibitem[Tu(2007)]{tu2007learning}
Zhuowen Tu.
\newblock Learning generative models via discriminative approaches.
\newblock In \emph{2007 IEEE Conference on Computer Vision and Pattern Recognition}, pages 1--8. IEEE, 2007.

\bibitem[Wang et~al.(2023)Wang, Li, Zhang, Xu, Zhou, Yu, Sheng, and Xu]{secret_segmenter}
Jinglong Wang, Xiawei Li, Jing Zhang, Qingyuan Xu, Qin Zhou, Qian Yu, Lu Sheng, and Dong Xu.
\newblock Diffusion model is secretly a training-free open vocabulary semantic segmenter.
\newblock \emph{arXiv preprint arXiv:2309.02773}, 2023.

\bibitem[Wu et~al.(2023)Wu, Zhao, Shou, Zhou, and Shen]{wu2023diffumask}
Weijia Wu, Yuzhong Zhao, Mike~Zheng Shou, Hong Zhou, and Chunhua Shen.
\newblock Diffumask: Synthesizing images with pixel-level annotations for semantic segmentation using diffusion models.
\newblock In \emph{Proceedings of the IEEE/CVF International Conference on Computer Vision}, pages 1206--1217, 2023.

\bibitem[Xiao et~al.(2023)Xiao, Yang, Zhou, and Zhang]{text_to_mask}
Changming Xiao, Qi Yang, Feng Zhou, and Changshui Zhang.
\newblock From text to mask: Localizing entities using the attention of text-to-image diffusion models.
\newblock \emph{arXiv preprint arXiv:2309.04109}, 2023.

\bibitem[Xu et~al.(2023{\natexlab{a}})Xu, Liu, Vahdat, Byeon, Wang, and De~Mello]{odise}
Jiarui Xu, Sifei Liu, Arash Vahdat, Wonmin Byeon, Xiaolong Wang, and Shalini De~Mello.
\newblock Open-vocabulary panoptic segmentation with text-to-image diffusion models.
\newblock In \emph{Proceedings of the IEEE/CVF Conference on Computer Vision and Pattern Recognition}, pages 2955--2966, 2023{\natexlab{a}}.

\bibitem[Xu et~al.(2023{\natexlab{b}})Xu, Liu, Wu, Tong, Li, Ding, Tang, and Dong]{imagereward}
Jiazheng Xu, Xiao Liu, Yuchen Wu, Yuxuan Tong, Qinkai Li, Ming Ding, Jie Tang, and Yuxiao Dong.
\newblock Imagereward: Learning and evaluating human preferences for text-to-image generation.
\newblock In \emph{Thirty-seventh Conference on Neural Information Processing Systems}, 2023{\natexlab{b}}.

\bibitem[Xue et~al.(2024)Xue, Song, Guo, Liu, Zong, Liu, and Luo]{xue2024raphael}
Zeyue Xue, Guanglu Song, Qiushan Guo, Boxiao Liu, Zhuofan Zong, Yu Liu, and Ping Luo.
\newblock Raphael: Text-to-image generation via large mixture of diffusion paths.
\newblock \emph{Advances in Neural Information Processing Systems}, 36, 2024.

\bibitem[Young et~al.(2014)Young, Lai, Hodosh, and Hockenmaier]{flickr30k_anno}
Peter Young, Alice Lai, Micah Hodosh, and Julia Hockenmaier.
\newblock From image descriptions to visual denotations: New similarity metrics for semantic inference over event descriptions.
\newblock \emph{Transactions of the Association for Computational Linguistics}, 2:\penalty0 67--78, 2014.

\bibitem[Yu et~al.(2022)Yu, Wang, Vasudevan, Yeung, Seyedhosseini, and Wu]{coca}
Jiahui Yu, Zirui Wang, Vijay Vasudevan, Legg Yeung, Mojtaba Seyedhosseini, and Yonghui Wu.
\newblock Coca: Contrastive captioners are image-text foundation models.
\newblock \emph{Transactions on Machine Learning Research}, 2022.

\bibitem[Zhang et~al.(2023{\natexlab{a}})Zhang, Rao, and Agrawala]{controlnet}
Lvmin Zhang, Anyi Rao, and Maneesh Agrawala.
\newblock Adding conditional control to text-to-image diffusion models.
\newblock In \emph{Proceedings of the IEEE/CVF International Conference on Computer Vision}, pages 3836--3847, 2023{\natexlab{a}}.

\bibitem[Zhang et~al.(2023{\natexlab{b}})Zhang, Zhang, Vineet, Joshi, and Wang]{control_gpt}
Tianjun Zhang, Yi Zhang, Vibhav Vineet, Neel Joshi, and Xin Wang.
\newblock Controllable text-to-image generation with gpt-4.
\newblock \emph{arXiv preprint arXiv:2305.18583}, 2023{\natexlab{b}}.

\bibitem[Zhao et~al.(2023)Zhao, Rao, Liu, Liu, Zhou, and Lu]{zhao2023unleashing}
Wenliang Zhao, Yongming Rao, Zuyan Liu, Benlin Liu, Jie Zhou, and Jiwen Lu.
\newblock Unleashing text-to-image diffusion models for visual perception.
\newblock In \emph{Proceedings of the IEEE/CVF International Conference on Computer Vision}, pages 5729--5739, 2023.

\bibitem[Zhou et~al.(2017)Zhou, Zhao, Puig, Fidler, Barriuso, and Torralba]{ade20k_scene}
Bolei Zhou, Hang Zhao, Xavier Puig, Sanja Fidler, Adela Barriuso, and Antonio Torralba.
\newblock Scene parsing through ade20k dataset.
\newblock In \emph{2017 IEEE Conference on Computer Vision and Pattern Recognition (CVPR)}, pages 5122--5130, 2017.

\bibitem[Zhou et~al.(2019)Zhou, Zhao, Puig, Xiao, Fidler, Barriuso, and Torralba]{ade20k_semantic}
Bolei Zhou, Hang Zhao, Xavier Puig, Tete Xiao, Sanja Fidler, Adela Barriuso, and Antonio Torralba.
\newblock Semantic understanding of scenes through the ade20k dataset.
\newblock \emph{International Journal of Computer Vision}, 127:\penalty0 302--321, 2019.

\bibitem[Zhou et~al.(2022)Zhou, Koltun, and Kr{\"a}henb{\"u}hl]{unidet}
Xingyi Zhou, Vladlen Koltun, and Philipp Kr{\"a}henb{\"u}hl.
\newblock Simple multi-dataset detection.
\newblock In \emph{Proceedings of the IEEE/CVF Conference on Computer Vision and Pattern Recognition}, pages 7571--7580, 2022.

\end{thebibliography}
}

\clearpage
\maketitlesupplementary

\section{Pipeline Setup}
\label{sec:pipeline_details}
We provide detailed data generation, training and inference settings for the \textsc{TokenCompose} pipeline setup in Table \ref{tab:pipeline_param}. Unless otherwise specified, experiments are run based on this set of settings. For settings that are not reported in Table \ref{tab:pipeline_param}, we follow the default values provided by their respective codebases.

\begin{table}[hb!]
\centering
\vspace{-2ex}
\scalebox{0.92}{
\begin{tabular}{l|l}
\hline
\toprule
\multicolumn{1}{l}{\multirow{-1}{*}{\textbf{Setting}}} & \textbf{Value} \\

\midrule
\textbf{\emph{Data Generation}} \\
\quad \small CLIP Model & \small \texttt{ViT-L/14@336px} \cite{clip, vit} \\
\quad \small POS Tagger & \small \texttt{flair/pos-english} \cite{flair} \\
\quad \small Target POS Tags & \small NN / NNS / NNP / NNPS \\
\quad \small G. DINO Model & \scriptsize \texttt{groundingdino\_swint\_ogc} \small \cite{grounding_dino, dino, swin} \\
\quad \small G. DINO Box Thres. & 0.25 \\
\quad \small G. DINO Text Thres. & 0.25 \\
\quad \small SAM Model & \small \texttt{sam\_hq\_vit\_h} \cite{vit, sam, sam_hq} \\

\textbf{\emph{Training}} \\
\quad \small Resolution \textsubscript{SD 1.4} & \small $512 \times 512$ \\
\quad \small Resolution \textsubscript{SD 2.1} & \small $768 \times 768$ \\
\quad \small Image Processing & \small Center Crop + Resize \\
\quad \small Batch Size & \small 1 \\
\quad \small Grad. Accum. Steps & \small 4 \\
\quad \small Grad. Ckpting. & \small \texttt{True} \\
\quad \small Train Steps \textsubscript{SD 1.4} & \small 24000 \\
\quad \small Train Steps \textsubscript{SD 2.1} & \small 32000 \\
\quad \small Learning Rate & \small 5e-6 \\
\quad \small LR Scheduler & \small Constant \\
\quad \small LR Warmup & \small \texttt{False} \\
\quad \small $\lambda$ for $\mathcal{L}_{token}$ & \small 1e-3 \\
\quad \small $\gamma$ for $\mathcal{L}_{pixel}$ & \small 5e-5 \\
\quad \small $\mathcal{L}_{token}$ and $\mathcal{L}_{pixel}$ & all layers in $U_{\text{Mid}}$ and $U_{\text{D}}$ \\
\quad \small Clf-Free Guidance \cite{classifier_free_guidance} & \small \texttt{False} \\

\textbf{\emph{Inference}} \\
\quad \small Resolution \textsubscript{SD 1.4} & \small $512 \times 512$ \\
\quad \small Resolution \textsubscript{SD 2.1} & \small $768 \times 768$ \\
\quad \small Timestep Scheduler & \small \texttt{PNDMScheduler} \\
\quad \small \# Inference Steps & \small 50 \\
\quad \small Clf-Free Guidance \cite{classifier_free_guidance} & \small \texttt{True} \\
\quad \small Guidance Scale & \small 7.5 \\

\bottomrule
\hline
\end{tabular}
}
\vspace{-2ex}
\caption{\textbf{Model choices and settings for data generation, training, and inference.} We provide a comprehensive list of pipeline details for data generation (\textit{e.g.,} caption selection, noun extraction and segmentation map generation), training (\textit{e.g.,} finetuning the Stable Diffusion model), and inference (\textit{e.g.,} evaluation of our finetuned Stable Diffusion model).}
\label{tab:pipeline_param}
\vspace{-4ex}
\end{table}

\section{Conditional Downstream Metrics}

\begin{figure}[ht!]
  \centering
    \includegraphics[width=1\linewidth]{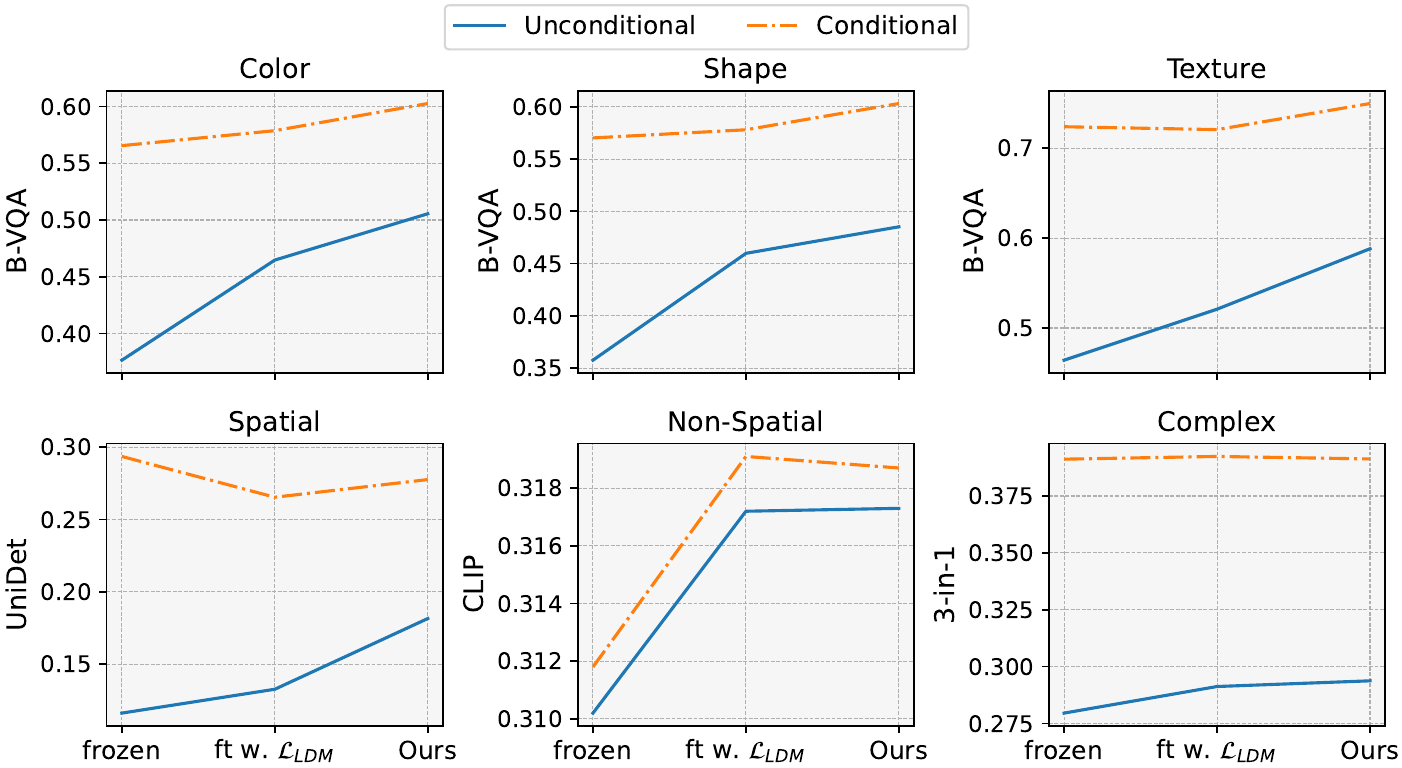}
    \vspace{-4ex}
    \caption{\textbf{Improvement comparison of unconditional and conditional downstream compositional metrics.} We illustrate that a significant margin of downstream compositional metrics is improved due to enhanced capabilities of multi-category instance composition. In this figure, we calculate the evaluation metrics from T2I-CompBench \cite{t2i_bench} by conditioning on the successful generation of all instances mentioned in the prompts, and compare the amount of quantitative improvement with the same metrics that are \emph{not} conditioned on successfull generation of all instances.}
    \label{fig:conditioned_metrics}
\vspace{-4ex}
\end{figure}

As multi-category instance composition serves as a prerequisite for successful downstream text-conditioned compositional generation, we conjecture that the improvements in downstream metrics are improved by higher chances of generating all instances mentioned in the prompt. In this section, we provide an additional object accuracy metric in Table \ref{tab:oa_t2i_bench}, and show that \textsc{TokenCompose} improves object accuracy in all downstream metrics being evaluated. Furthermore, in Figure \ref{fig:conditioned_metrics}, we demonstrate how better multi-category instance composition capabilities can improve these metrics by plotting the improvement curve from \textbf{\emph{(1)}} a frozen Stable Diffusion to \textbf{\emph{(2)}} a Stable Diffusion finetuned with only $\lsimplelcm$ objective, and finally to \textbf{\emph{(3)}} a Stable Diffusion model finetuned with both $\lsimplelcm$ and our grounding objectives conditioned and unconditioned on successful generation of all instances from the compositional prompts. 

From the results, we observe that the improvements in all attribute binding metrics (\textit{e.g.,} color, shape, texture) and the majority of object relation metrics (\textit{e.g.,} spatial, complex) are much more significant in the unconditional case than in the conditional case. The only downstream benchmark where the difference in improvement is insignificant is the \emph{non-spatial} compositionality. We believe that this insignificance can be explained by two factors: (1) lowest amount of improvement in object accuracy for this specific downstream benchmark, as shown in Table \ref{tab:oa_t2i_bench} and (2) relatively low correlation between automatic scores (\textit{i.e.,} CLIP Score \cite{clip, clip_score}) and human ratings among all compositional benchmarks from the T2I-CompBench \cite{t2i_bench}; this indicates that the evaluation model may have a comparably weak discriminative capability for this specific task.

\begin{table}[ht]
\centering
\scalebox{0.75}{
\begin{tabular}{l|cccccc}
\hline
\toprule
\multicolumn{1}{l}{\multirow{-1}{*}{\textbf{Model}}} & \small Color & \small Shape & \small Texture & \small Spatial & \small Non-Spat. & \small Complex \\

\midrule
\textbf{\emph{SD 1.4}} \\
\quad frozen & 47.40 & 25.33 & 15.27 & 19.40 & 45.10 & 26.60  \\
\quad ft. w. \small $\lsimplelcm$ & 55.70 & 28.49 & 18.95 & 27.05 & 47.33 & 29.33 \\
\midrule
\colorcello \quad \emph{\textbf{Ours}} & \colorcello \textbf{62.92} & \colorcello \textbf{32.95} & \colorcello \textbf{25.59} & \colorcello \textbf{33.30} & \colorcello \textbf{48.94} & \colorcello \textbf{32.10} \\

\bottomrule
\hline
\end{tabular}
}
\vspace{-2ex}
\caption{\textbf{Object accuracy in downstream compositional metrics.} We calculate the object accuracy metric (\textit{i.e.,} success rate of generating all instances in the prompt based on a detection model \cite{owlv2}) for each of the compositional benchmarks from T2I-CompBench \cite{t2i_bench}.}
\label{tab:oa_t2i_bench}
\vspace{-2ex}
\end{table}

\begin{figure}[ht]
  \centering
  \vspace{-0.5ex}
    \includegraphics[width=1\linewidth]{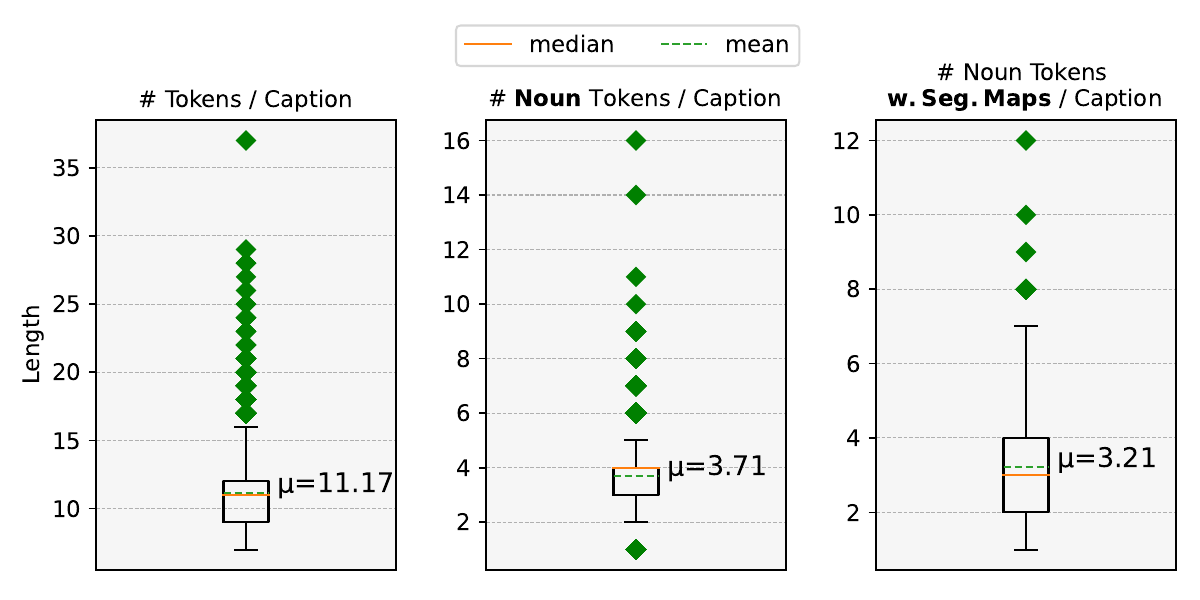}
    \vspace{-4ex}
    \caption{\textbf{Approximations of the average number of tokens with grounding objectives per training prompt.} From left to right, we show the distribution of number of tokens per caption, number of noun tokens per caption, and number of noun tokens that have generated segmentation maps per caption.}
    \label{fig:dataset_metadata}
\vspace{-3ex}
\end{figure}

\begin{figure*}[ht]
  \centering
  \vspace{-2ex}
  \includegraphics[width=1\linewidth]{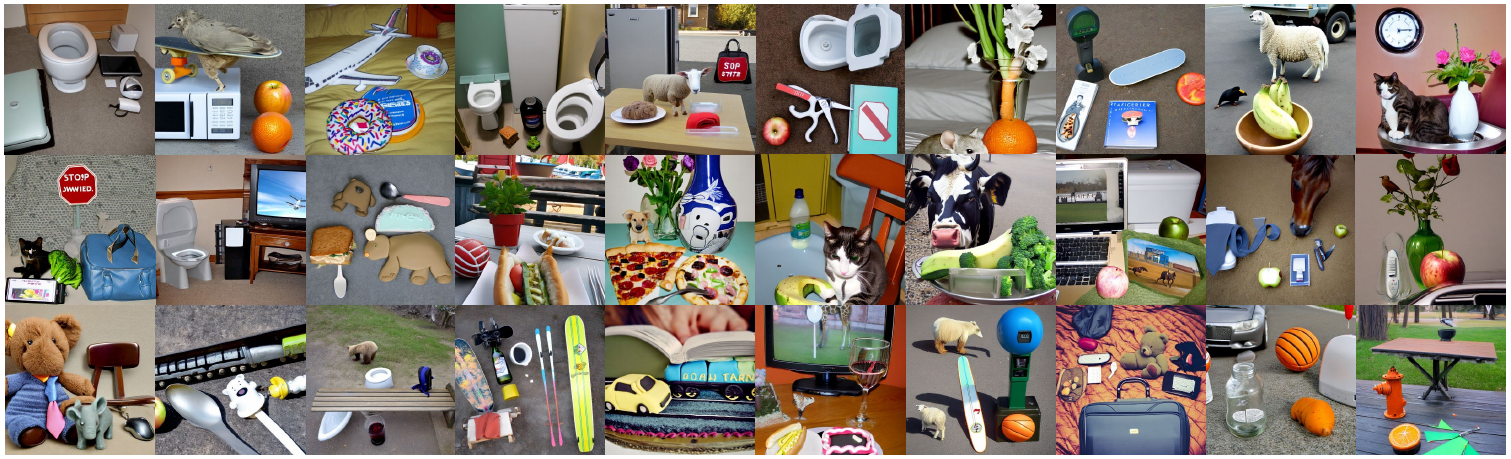}
  \vspace{-4ex}
   \caption{\textbf{More samples} in multi-category instance composition.}
   \label{fig:phys_rules_aggr}
\end{figure*}

\begin{figure*}[ht]
  \centering
  \vspace{-2ex}
  \includegraphics[width=1\linewidth]{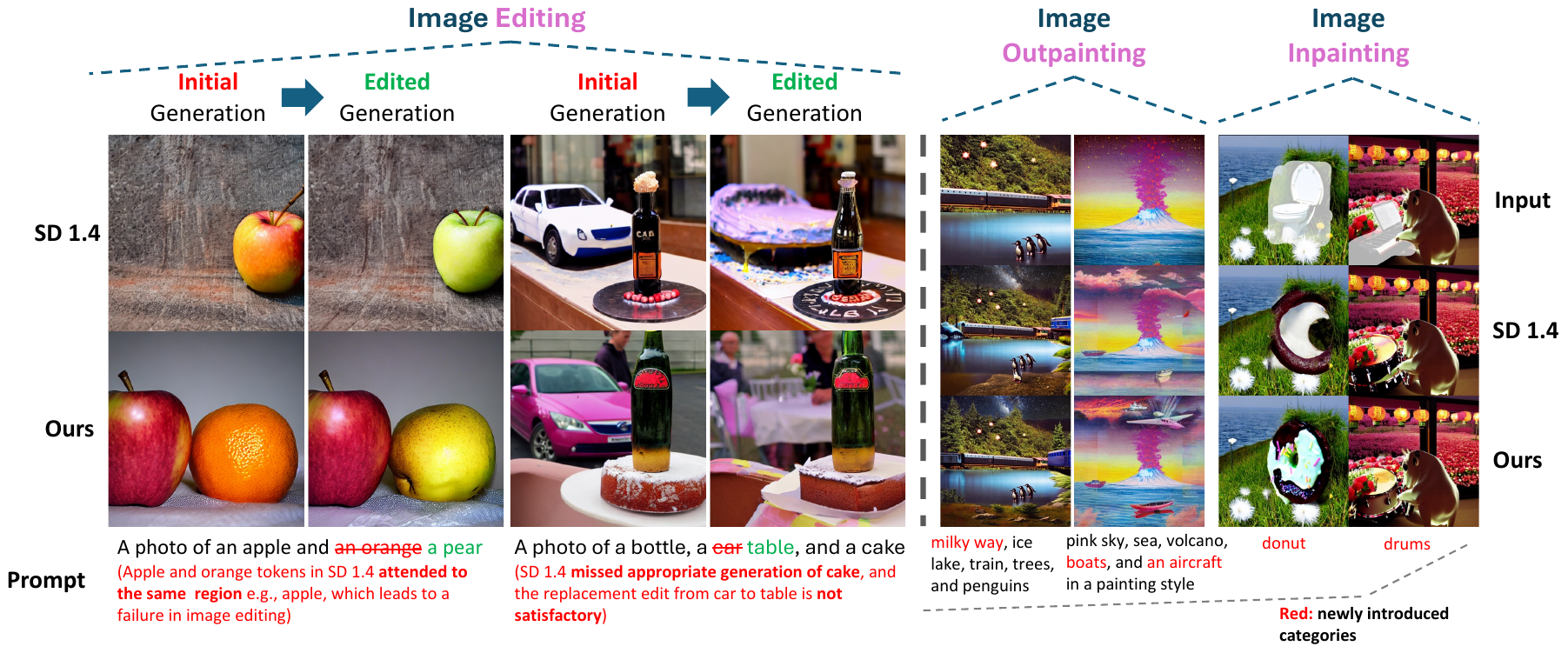}
  \vspace{-5ex}
   \caption{\textbf{Downstream applications} in prompt-to-prompt \cite{hertz2022prompt} image editing and zero-shot outpainting and inpainting.}
   \label{fig:downstream_apps}
\end{figure*}

\begin{figure*}[ht]
  \centering
  \includegraphics[width=1.0\linewidth]{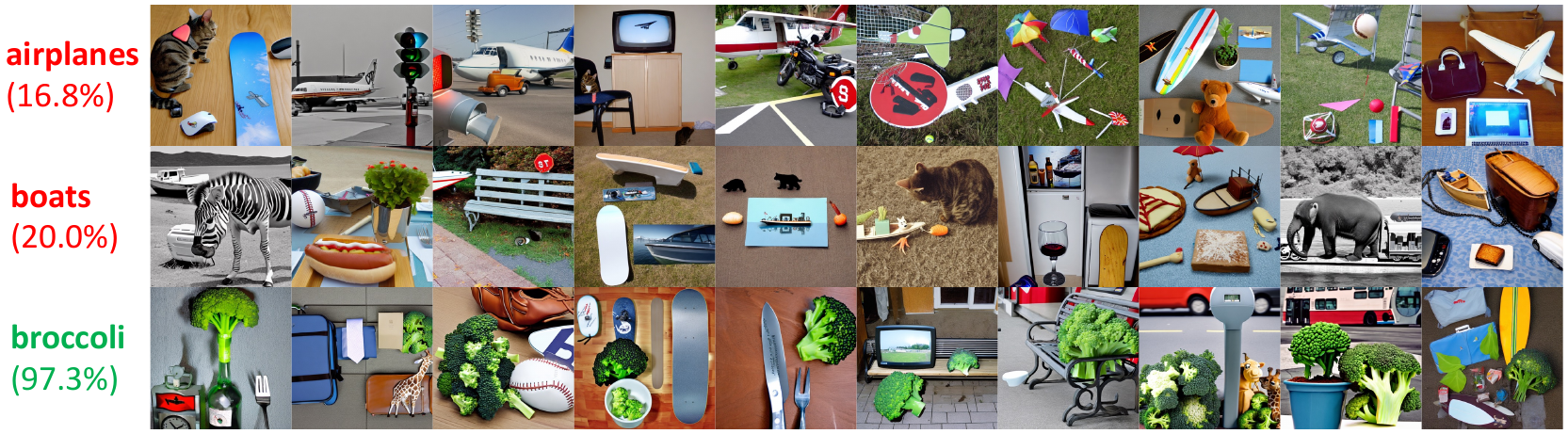}
  \vspace{-5ex}
   \caption{Categories with poor \& strong performance}.
   \label{fig:succ_rate_exp}
   \vspace{-2ex}
\end{figure*}

\begin{figure*}[ht]
  \centering
  \includegraphics[width=1.0\linewidth]{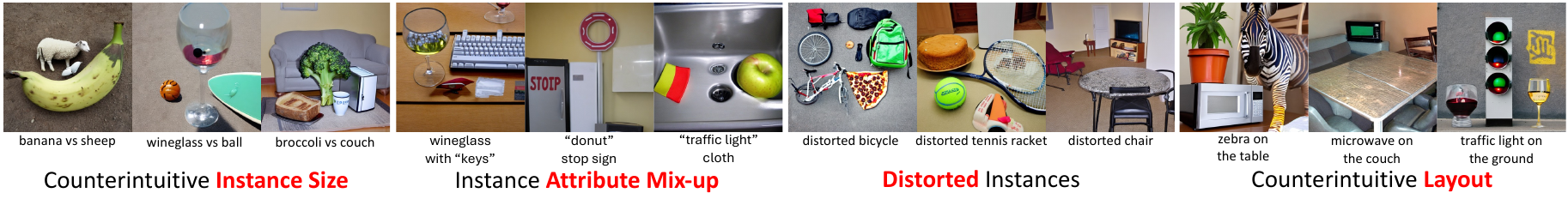}
  \vspace{-5ex}
   \caption{\textbf{Failure case analysis} in multi-category compositionality.}
   \label{fig:fail_case_analysis}
   \vspace{-2ex}
\end{figure*}

\begin{figure*}[ht!]
  \centering
    \includegraphics[width=1\linewidth]{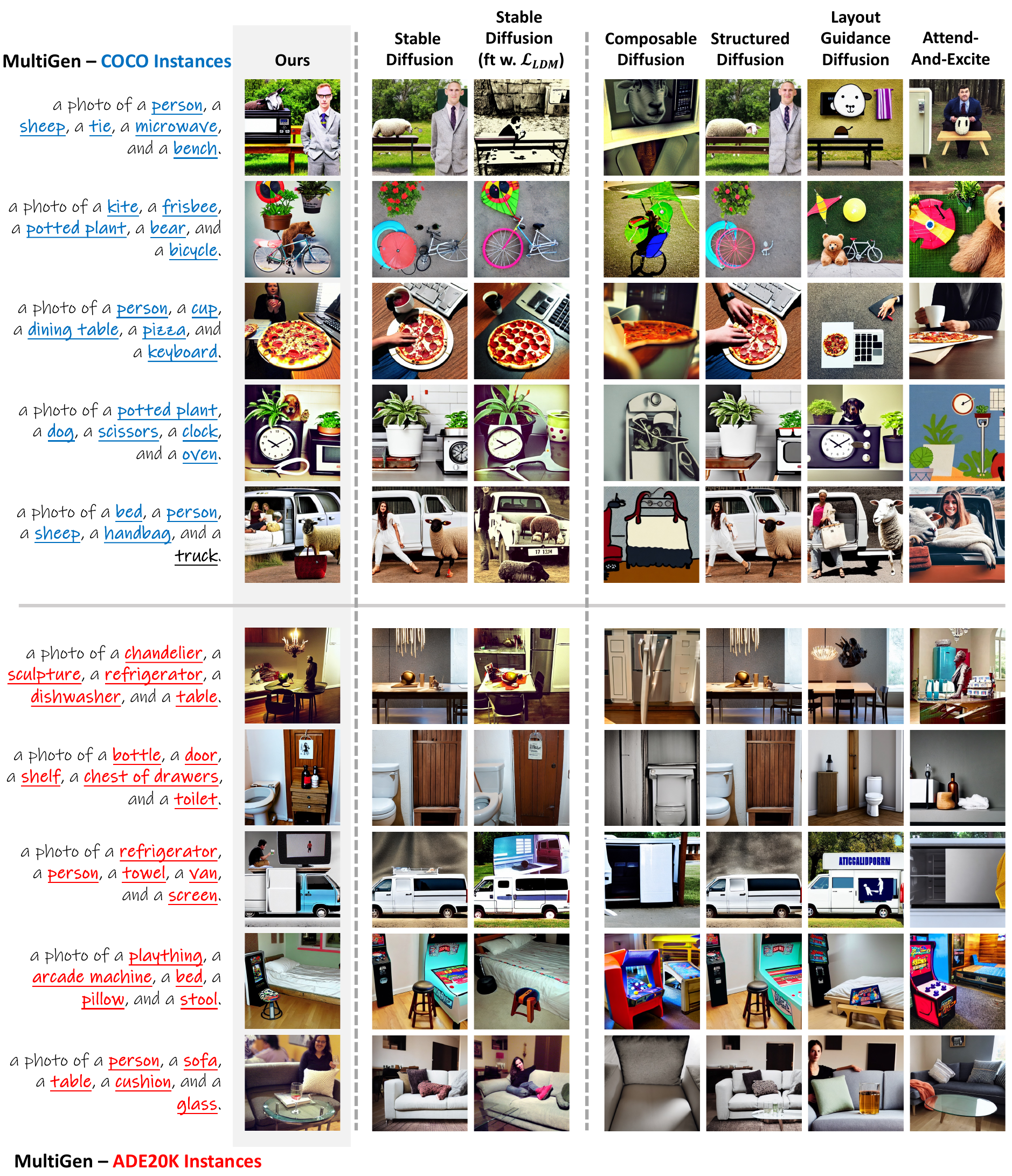}
    \caption{\textbf{Sampled images with prompts from our \textsc{MultiGen} benchmark.} To facilitate understanding of our multi-category instance composition benchmark as well as further qualitative comparisons among different baselines, we provide sampled images from \textsc{MultiGen} with COCO instances as well as ADE20K instances.}
    \label{fig:multigen_viz}
\vspace{-4ex}
\end{figure*}

\begin{figure*}[ht!]
  \centering
    \includegraphics[width=1\linewidth]{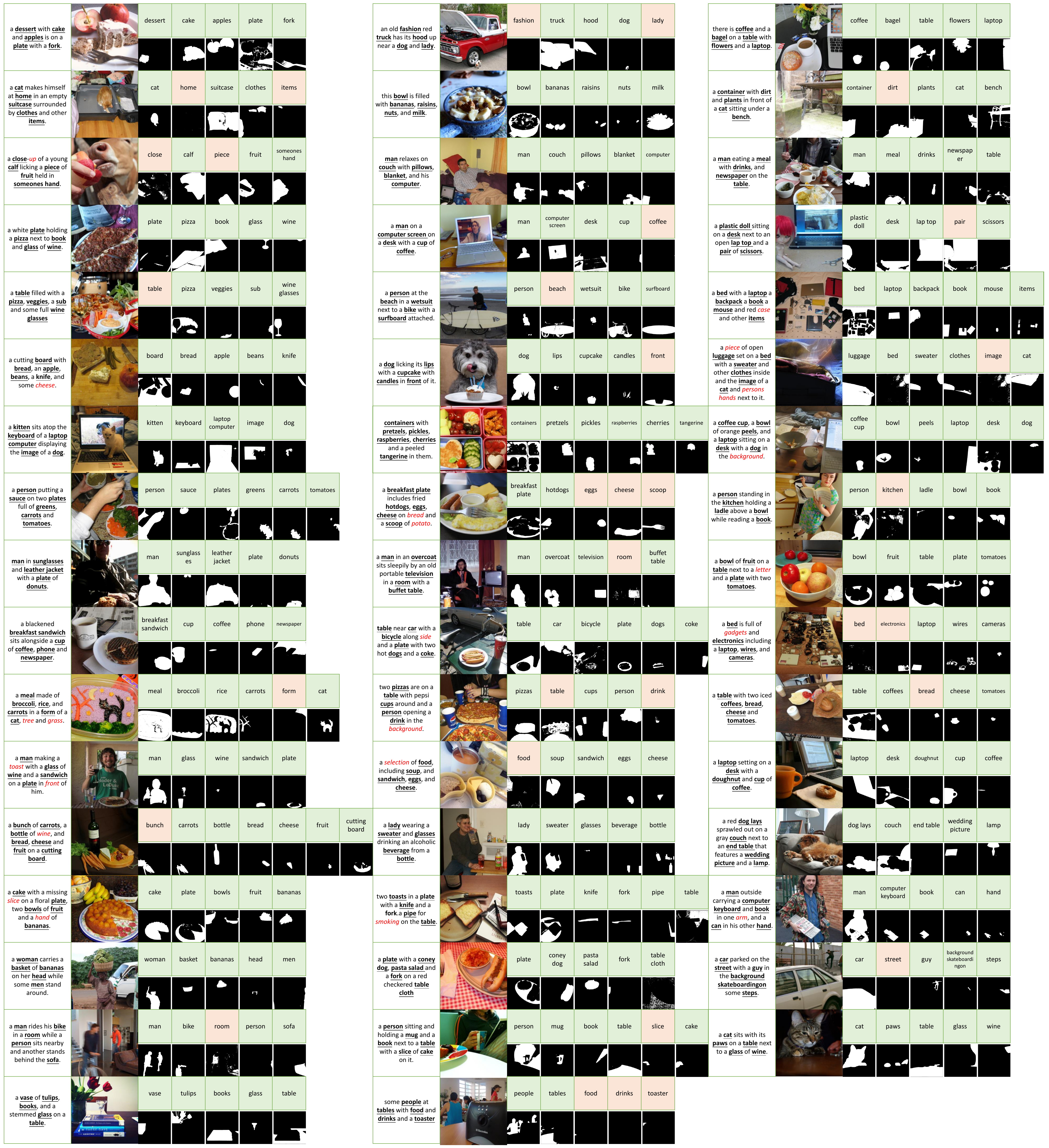}
    \caption{\textbf{Grounding dataset visualization.} We provide visualizations of 50 selected image-text pairs (number of noun tokens with segmentation maps $\geq 5$) and their corresponding token-level binary segmentation maps from the training data to facilitate understanding of our training dataset.}
    \label{fig:dataset_vis}
\vspace{-4ex}
\end{figure*}

\section{More Examples}
\textbf{Multi-category Instance Composition.} We provide more visualizations in Figure. \ref{fig:phys_rules_aggr} to illustrate capabilities of TokenCompose in multi-category instance composition. In addition to being able to generate multiple categories of instances successfully, we show that object affordance (\textit{e.g.,} attach, sit, support, \textit{etc}) can be maintained, which indicates that TokenCompose is able to implicitly learn the basic ``physical rules" via the object token and segmentation consistency constraints. 

\textbf{Downstream Applications.} We show in Figure. \ref{fig:downstream_apps} positive impacts in downstream applications. As expected, TokenCompose shows its effectiveness in tasks beyond text-to-image generation.

\textbf{Benchmark.} \textsc{MultiGen} benchmark plays a crucial role in evaluating multi-category instance composition capabilities of text-to-image models. To offer a more intuitive sense in this compositional task, we visualize the generated images from various prompts in this benchmark (\textit{e.g.,} MG COCO Instances and MG ADE20K Instances). We use the same latent for each comparison for fairness. The images are presented in Figure \ref{fig:multigen_viz}.

\section{Grounded COCO Dataset}
As seen in Figure \ref{fig:method}, we adopt a POS Tagger and the Grounded SAM \cite{flair, grounding_dino, sam} to extract binary segmentation maps from noun tokens from the image-text pair dataset. We aim to expand the visualizations of the generated data in Figure \ref{fig:dataset_vis}. Each row and column represents a single data that the model is trained with. From the left to the right of the data are the caption, the input image, and the grounded binary segmentation maps. 

The bold and underlined \underline{\textbf{text}} in the caption represents noun tokens captured by the POS tagger where their corresponding segmentation maps are extracted from the Grounded SAM. Italicized and red \textcolor{red}{\textit{text}} in the caption represents noun tokens captured by the POS tagger but does \emph{not} have segmentation maps extracted from the Grounded SAM due to the model not being able to locate the objects.

Grounded segmentation maps on the right are paired with their correpsonding tokens. For tokens with a green \colorbox{sp_lightgreen}{background}, their segmentation maps successfully capture the aligned contents in the image, whereas for a small fraction of tokens with an orange \colorbox{sp_lightorange}{background}, their segmentation maps do not capture the aligned contents in the image.

We also provide reference on approximations (\textit{i.e.,} word split by space) of (1) average number of tokens per caption, (2) average number of \emph{noun tokens} per caption, and (3) average number of noun tokens that \emph{have their corresponding segmentation maps} per caption in Figure \ref{fig:dataset_metadata}. As shown in the figure, our training dataset contains an average of 3.71 noun tokens overall and 3.21 noun tokens that have their corresponding segmentation maps.

\section{Analysis}

\begin{figure*}[ht!]
  \centering
    \includegraphics[width=0.95\linewidth]{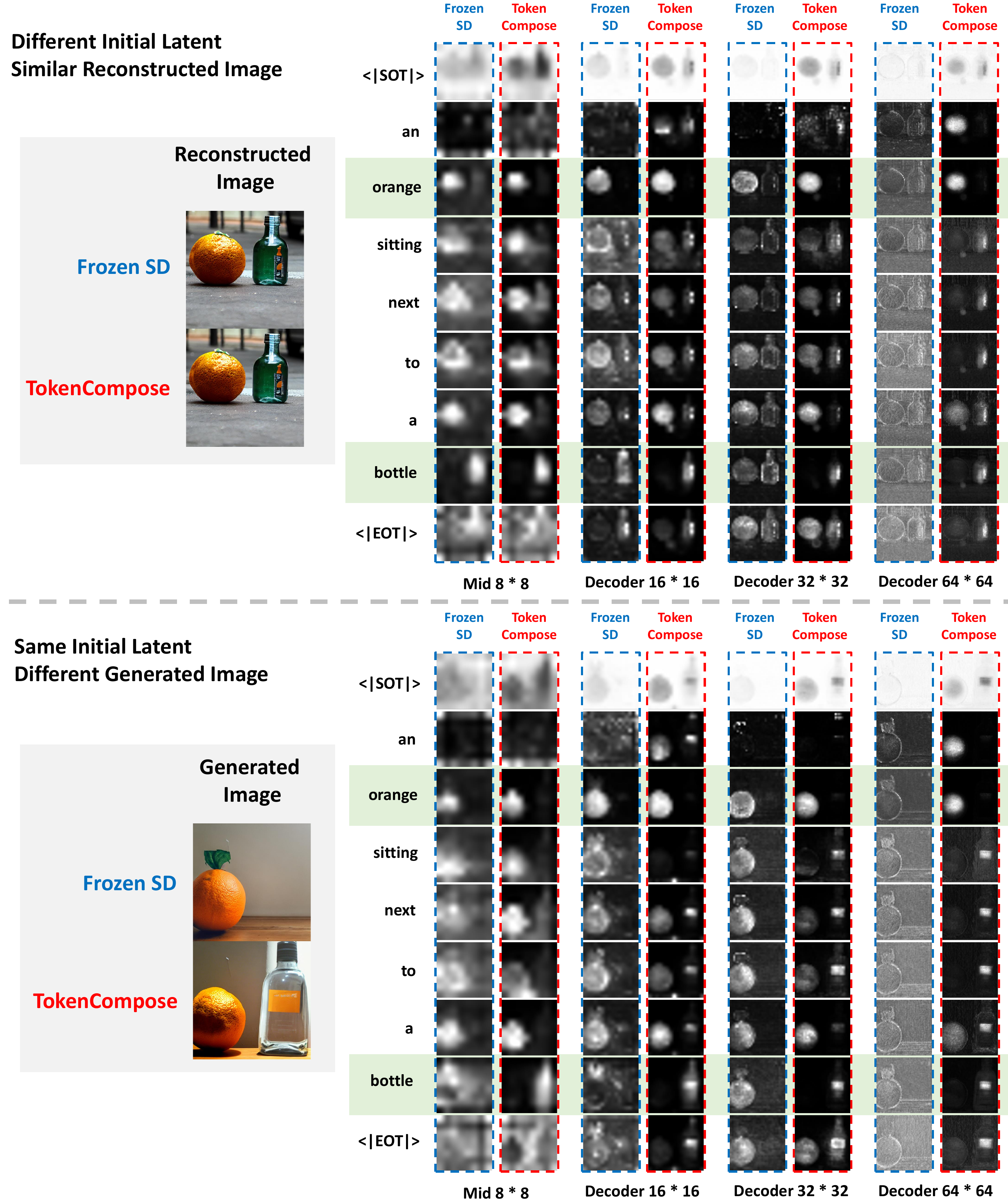}
    \caption{\textbf{Cross-attention map visualizations by different cross-attention layers at denoising U-Net.} We visualize the cross-attention map between a frozen Stable Diffusion \cite{ldm} model and our model at the middle block and decoder layers of the denoising U-Net, where cross-attention at these layers are trained with our grounding objectives. In the upper example, we leverage null text inversion \cite{null_text_inversion} to let two models reconstruct similar images using different latents for comparable cross-attention maps to demonstrate stronger grounding capabilities of our model. In the lower example, we use the same initial latent for two different models to generate images to demonstrate how stronger grounding capabilities lead to better compositionality.}
    \label{fig:ca_by_res}
\vspace{-4ex}
\end{figure*}

\begin{figure*}[ht!]
  \centering
    \includegraphics[width=1\linewidth]{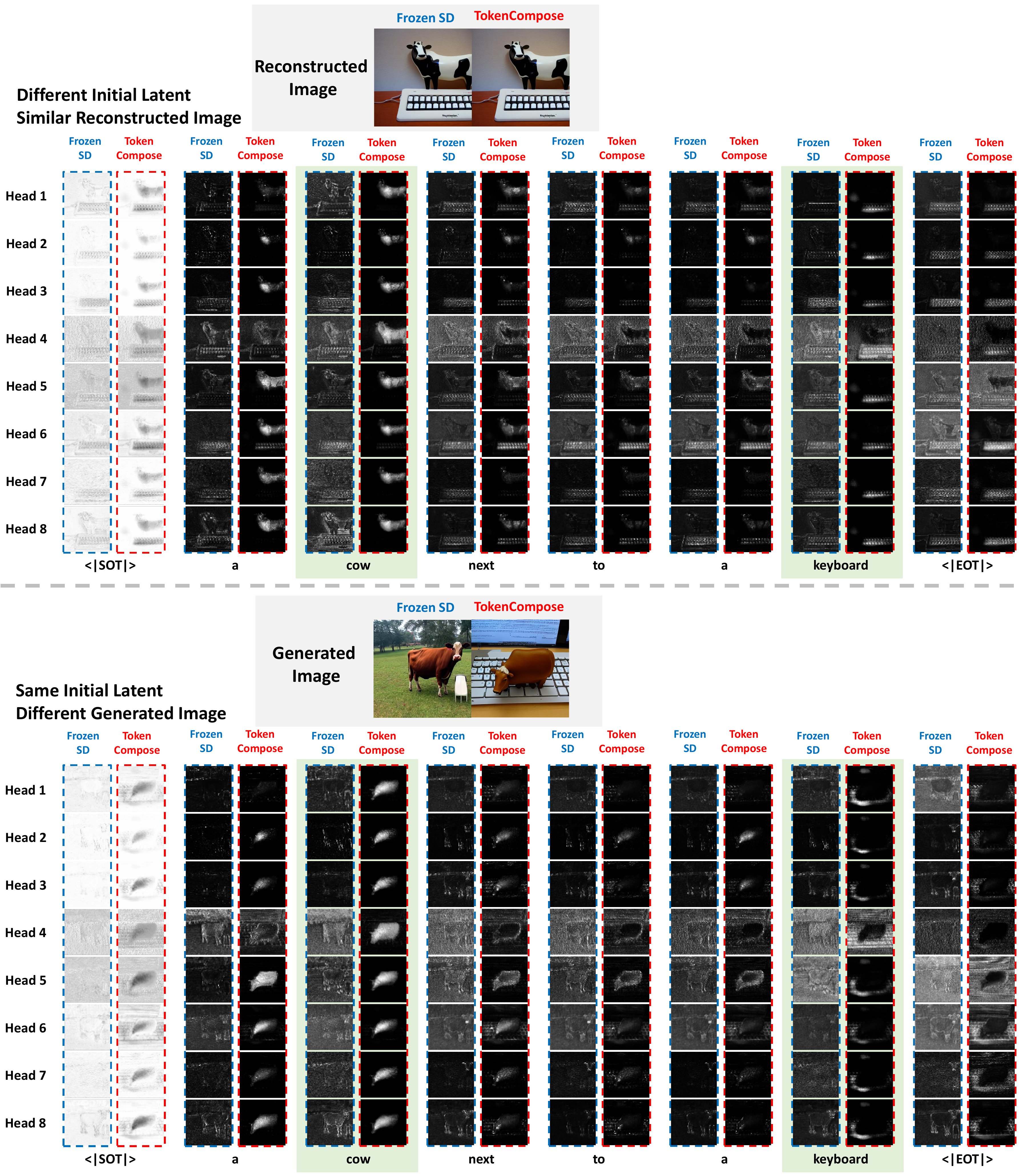}
    \caption{\textbf{Cross-attention map visualizations by different cross-attention heads at the denoising U-Net.} We visualize the cross-attention map of each head at the last layer (\textit{i.e.,} $U_{\text{D}}^{64 \times 64}$) of the U-Net decoder between a frozen Stable Diffusion \cite{ldm} model and our model. We use the same setting as in Figure \ref{fig:ca_by_res} but with a different prompt for a more diverse visualization. We show that our grounding objective \emph{allows} flexibility of different heads to attend to different regions of the latent.}
    \label{fig:ca_by_head}
\vspace{-4ex}
\end{figure*}

\begin{figure*}[ht!]
  \centering
    \includegraphics[width=0.9\linewidth]{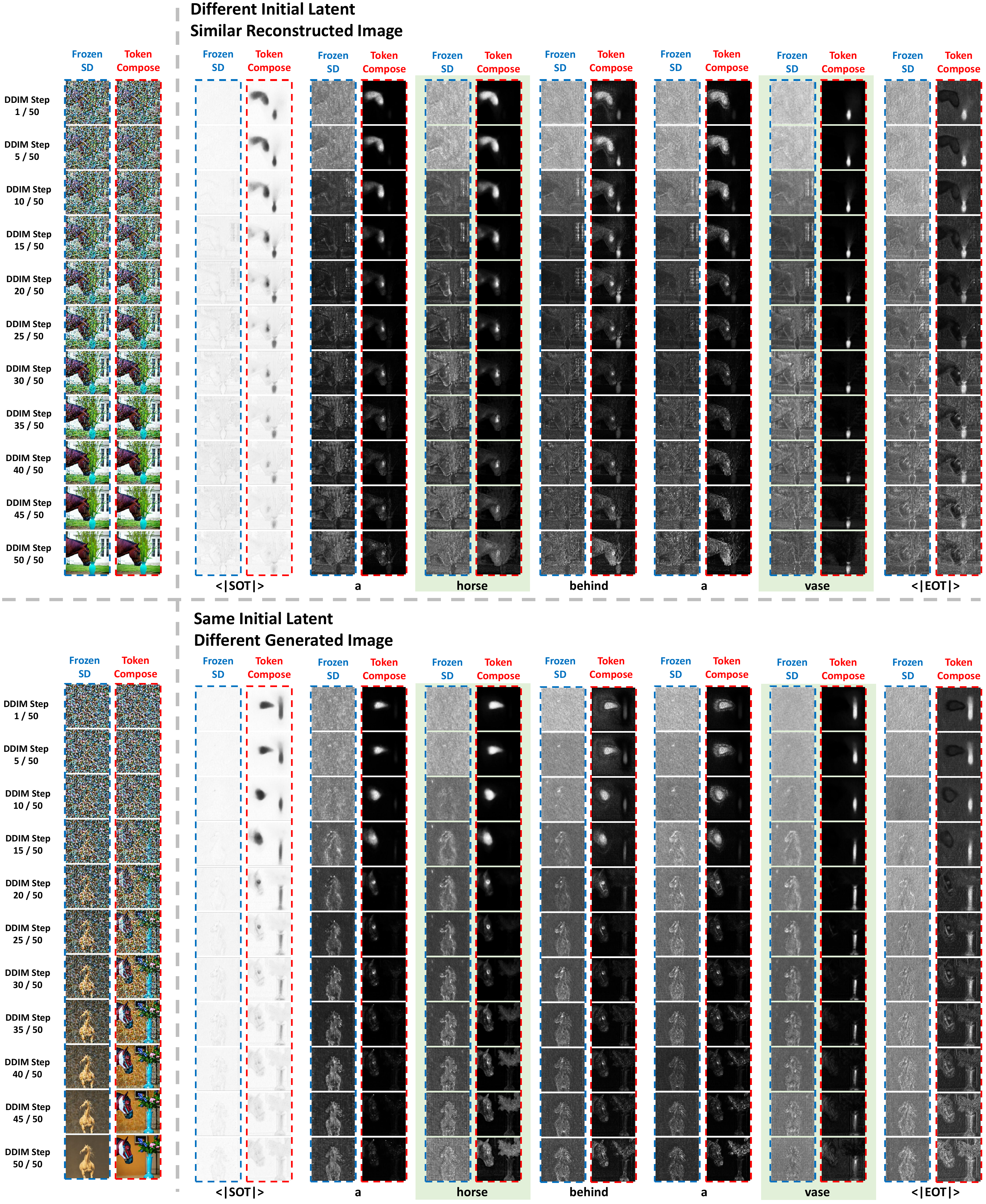}
    \caption{\textbf{Cross-attention map visualizations by different denoising time steps at the denoising U-Net.} We visualize the cross-attention map of the last layer (\textit{i.e.,} $U_{\text{D}}^{64 \times 64}$) of the U-Net decoder between a frozen Stable Diffusion \cite{ldm} model and our model at time step 1 and every 5 steps based on a 50-step DDIM \cite{ddim} scheduler. We use the same setting as in Figure \ref{fig:ca_by_res} with a different prompt for a more diverse visualization. We show that our grounding objective enables cross-attention of different object tokens to aggregate at different regions of the noisy latent \emph{early} during inference. This enables the model to generate different categories of instances more successfully, leading to better multi-category instance composition capabilities.}
    \label{fig:ca_by_timestep}
\vspace{-4ex}
\end{figure*}

\textbf{Attention Visualizations.} To gain a better understanding of how incorporating grounding objectives into text-to-image models during training affects cross-attention maps for image reconstruction \cite{null_text_inversion} \& generation tasks at inference time, we provide token-level cross-attention map visualizations on three axes: \emph{\textbf{(1)}} different cross-attention layers with various resolutions (Figure \ref{fig:ca_by_res}); \emph{\textbf{(2)}} different heads of the multi-head cross-attention (Figure \ref{fig:ca_by_head}; and \emph{\textbf{(3)}} different timesteps during the denoising process (Figure \ref{fig:ca_by_timestep}).

\begin{figure*}[ht!]
  \centering
    \includegraphics[width=1\linewidth]{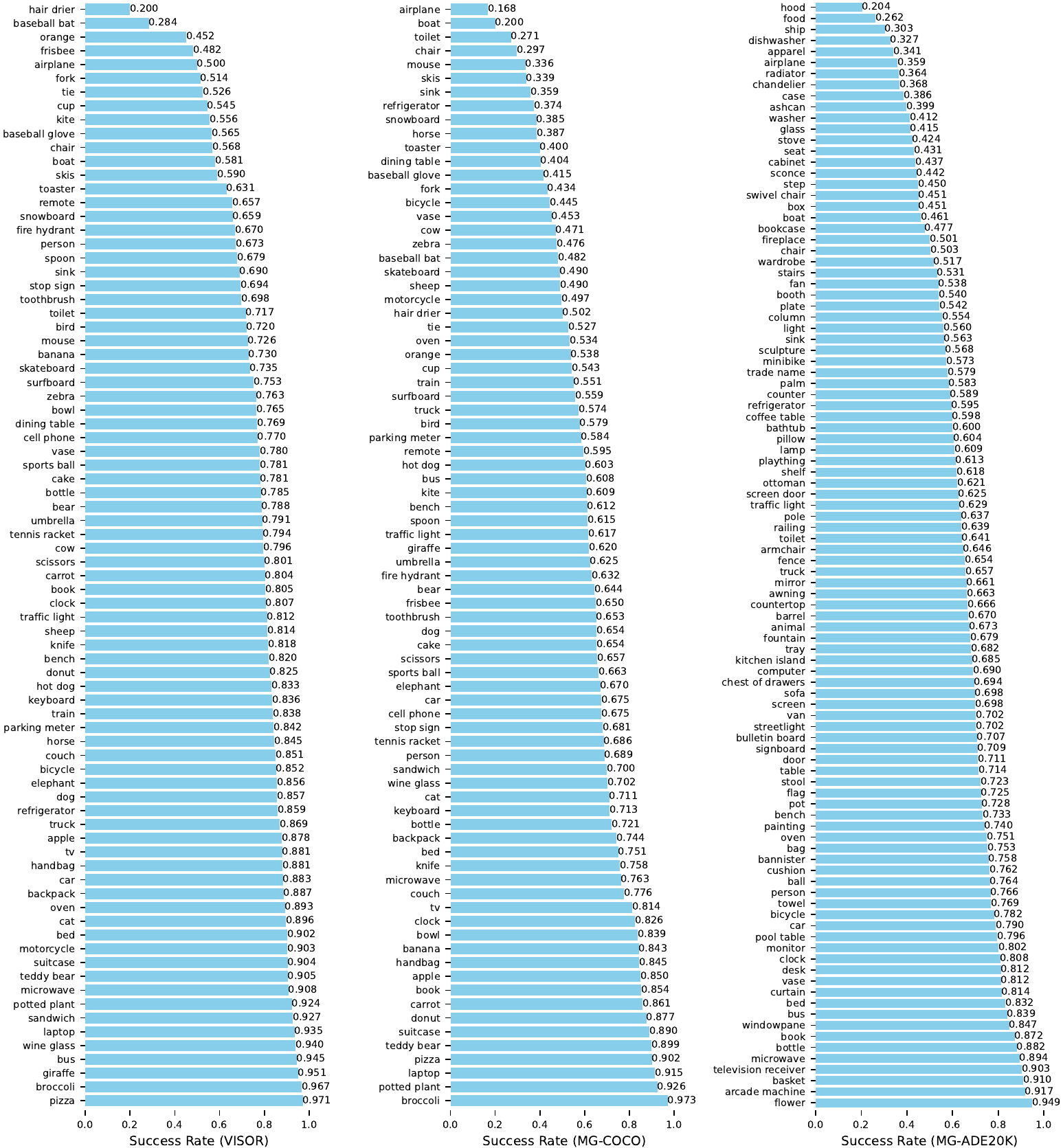}
    \caption{\textbf{Instance generation success rate in multi-category instance composition benchmarks.} We provide the success rate (from a 0-1 scale) of generating instances with our best-performing model for VISOR \cite{visor}, MG-COCO, and MG-ADE20K benchmarks.}
    \label{fig:success_rate}
\vspace{-4ex}
\end{figure*}

\textbf{Multi-category Instance Composition Success Rate.} 
Given a set of compositional prompts, we calculate the count of each category that appears in these prompts along with images where the instance(s) of this category is/are detected by a detection model \cite{owlv2}. We divide the number of images that contain a specific category of the instance by the number of prompts that contains this category to acquire the success rate and report the numbers in Figure \ref{fig:success_rate}.

\textbf{Failure Cases.} 
We provide generated and detected samples for categories with low and high success rates in Figure. \ref{fig:succ_rate_exp}. We believe that one explanation for the poor performance categories have variants and viewpoints with drastic visual differences, so learning and generating them is harder. Further, we aggregate patterns of common failures in our multi-category instance composition in Figure. \ref{fig:fail_case_analysis}. We believe that visual commonsense reasoning aspect of the generative model would be an area of improvement.

\end{document}